\documentclass{article}

\usepackage{arxiv}

\usepackage[utf8]{inputenc} 
\usepackage[T1]{fontenc}    
\usepackage{hyperref}       
\usepackage{url}            
\usepackage{booktabs}       
\usepackage{nicefrac}       
\usepackage{microtype}      
\usepackage{lipsum}		
\usepackage{graphicx}
\usepackage{natbib}
\usepackage{doi}

\usepackage{multirow}%
\usepackage{amsmath,amssymb,amsfonts}%
\usepackage{amsthm}%
\usepackage{mathrsfs}%
\usepackage[title]{appendix}%
\usepackage{xcolor}%
\usepackage{textcomp}%
\usepackage{manyfoot}%
\usepackage{algorithm}%
\usepackage{algorithmicx}%
\usepackage{algpseudocode}%
\usepackage{listings}%
\usepackage{fullpage}%
\usepackage{setspace}%

\title{Contrastive Learning of English Language and Crystal Graphs for Multimodal Representation of Materials Knowledge}

\author{{\hspace{1mm}Yang Jeong Park} \\
    Massachusetts Institute of Technology\\
    Cambridge, MA 02139, USA \\
    \texttt{parkyj@mit.edu} \\
    \And{\hspace{1mm}Mayank Kumaran} \\
    University of Illinois Urbana-Champaign\\
    Urbana, IL 61801, USA \\
    \texttt{mayank6@illinois.edu} \\
    \And{\hspace{1mm}Chia-Wei Hsu} \\
    Massachusetts Institute of Technology\\
    Cambridge, MA 02139, USA \\
    \texttt{acc1107@mit.edu} \\
    \And{\hspace{1mm}Elsa Olivetti}\thanks{Corresponding authors: \texttt{elsao@mit.edu}, \texttt{liju@mit.edu}} \\
    Massachusetts Institute of Technology\\
    Cambridge, MA 02142, USA \\
    \texttt{elsao@mit.edu} \\
    \And{\hspace{1mm}Ju Li*} \\
    Massachusetts Institute of Technology\\
    Cambridge, MA 02139, USA \\
    \texttt{liju@mit.edu} \\
}

\date{}

\renewcommand{\shorttitle}{\textit{arXiv} Template}

\hypersetup{
pdftitle={A template for the arxiv style},
pdfsubject={q-bio.NC, q-bio.QM},
pdfauthor={David S.~Hippocampus, Elias D.~Striatum},
pdfkeywords={First keyword, Second keyword, More},
}

\begin{document}
\maketitle

\begin{abstract}
Artificial intelligence (AI) is increasingly used for the inverse design of materials, such as crystals and molecules. Existing AI research on molecules has integrated chemical structures of molecules with textual knowledge to adapt to complex instructions. However, this approach has been unattainable for crystals due to data scarcity from the biased distribution of investigated crystals and the lack of semantic supervision in peer-reviewed literature. In this work, we introduce a contrastive language-crystals model (CLaC) pre-trained on a newly synthesized dataset of 126k crystal structure-text pairs. To demonstrate the advantage of using synthetic data to overcome data scarcity, we constructed a comparable dataset extracted from academic papers. We evaluate CLaC's generalization ability through various zero-shot cross-modal tasks and downstream applications. In experiments, CLaC achieves state-of-the-art zero-shot generalization performance in understanding crystal structures, surpassing latest large language models.
\end{abstract}


\section{Introduction}

Discovering new materials that push the performance limits for certain applications is the ultimate goal for most materials scientists. To accelerate crystal discovery, several applications of machine learning (ML), including virtual screening \cite{kim2017VirtualScreening, sorkun2020AI-VirtualScreening}, property prediction \cite{xie2018CGCNN, choudhary2021ALIGNN, kaundinya2022predictionALIGNN}, generative models \cite{hoffmann2019Voxel-VAE, noh2019inverseMatter, court2020Cond-DFC-VAE, xie2021CDVAE, lyngby2022CDVAE2Ddiscovery, gong2023GNNforcrystal, pakornchote2024dpcdvae}, and active learning \cite{lookman2019ALinMaterials, chen2020GenerativeInverseDesignUsingAL}, have been explored. Recent ML applications mainly focus on modeling 3D crystal graphs \cite{choudhary2021ALIGNN, kaundinya2022predictionALIGNN} via graph neural networks (GNNs) to reduce dependence on hand-crafted features. 

However, the limited availability of labels, due to labor-intensive experiments and computation-intensive quantum chemical calculations, hinders the generalization to unseen tasks \cite{butler2018MLforMolecularandMaterialsScience, agrawal2016FourthParadigmPerspective}. Self-supervised learning techniques are proposed to circumvent costly labeling processes and have been applied across various domains, including natural language\cite{devlin2018BERT, radford2019GPT2}, images\cite{kolesnikov2019SSLVision}, biochemical molecules\cite{wang2022MolCLR, fang2022GEM} and crystals\cite{koker2022CrystalCLR, magar2022CrystalTwins}. Among these approaches, contrastive learning, which aims to maintain representations of positive pairs closely in latent space while separating negative pairs, has proven effective in generalizing to a wide range of unseen downstream tasks across multiple modalities. For instance, contrastive language-image pre-training (CLIP) \cite{radford2021CLIP} serves as a foundational building block for text-based image generation models, such as Stable Diffusion\cite{rombach2022StableDiffusion} and DALL-E\cite{ramesh2021Dalle}, as well as vision-language models like LLaVA\cite{liu2023llava, li2023llava-med} for open-ended question answering. 

Despite the advantages of multimodal representations, multimodal learning for crystal design is less explored due to unique challenges. The first challenge is data scarcity. For example, CLIP was trained on 400 million image-caption pairs in the computer vision field, and in the molecular domain, more than 1 billion molecules are accessible in ZINC database\cite{irwin2005zinc}. In contrast, the amount of crystal data available is limited to just a few million \cite{belsky2002ICSD, jain2013MaterialsProject}. The second issue is the biased data distribution, similar to the ``income inequality'' problem in econometrics. The research efforts undertaken by human researchers tend to be heavily focused on a very small set of crystals, the so-called ``hot'' materials. This trend leads to gross neglect of the huge number of other possibilities on the thermodynamic convex hull, e.g. those possibilities allowed by first-principles considerations alone without the human selection bias. These sociological factors pose a challenge to ensuring sufficient diversity in datasets of natural language descriptions that explain their physical properties. In contrast, existing unimodal deep learning research for materials science, including GNNs, can easily access a vast and diverse array of millions of virtual crystal data. Data complexity is the third. Unlike molecules with lower degrees of freedom in constituent elements such as hydrogen and carbon, crystals exhibit much higher degrees of freedom in their constituent elements despite less data. These challenges have prevented previous research from developing multimodal learning systems that combine crystalline materials and natural language, leading to a lack of standard metrics or benchmarks necessary to evaluate them.

In a previous study by Park et al. \cite{park2023materialsnarratives}, it was proposed to address the data scarcity issue by generating GPT-synthesized narrative data using chatbots. Building on this work, we introduce the Contrastive Language-Crystal model (CLaC), a framework designed to simultaneously learn the 3D atomic structure of crystals and their linguistic representations. To demonstrate the effect of multimodal contrastive learning, we designed evaluation tests such as zero-shot information retrieval, named entity recognition (NER), and paper abstract classification (PAC). To highlight the importance of synthetic data, we also trained the same model on a corpus extracted from materials science academic literature for comparison. Empirically, our CLaC model achieves the best performance on zero-shot information retrieval. We also show that transfer learning using pre-trained multimodal contrastive models outperforms unimodal models in NER and PAC tasks.

\section{Related Work}
\paragraph{Self-supervised learning for molecules and crystals}

GNNs have significantly impacted material property predictions \cite{xie2018CGCNN, gilmer2017MPNN} by providing an alternative way for complex quantum chemical computations. However, the costly labeling process requiring expensive quantum chemical calculations is the bottleneck in GNN application. To address this challenge, various self-supervised learning (SSL) methods have been proposed. Wang et al. \cite{wang2022MolCLR} reduced the need for expensive quantum chemical labeling by applying contrastive learning to unlabeled 10 million molecules using molecular GNNs. Koker et al. \cite{koker2022CrystalCLR} emphasized the importance of augmentation techniques when applying contrastive learning to encoding periodic crystals as graphs with CGCNNs as encoders. Magar et al. \cite{magar2022CrystalTwins} improved the performance of CGCNNs by applying Barlow twins \cite{zbontar2021barlowtwins}, a self-supervised learning method that does not require negative pairs.

\paragraph{Multimodal learning for molecules}
In the molecular domain, performing multimodal contrastive learning between graphs and text has been demonstrated to improve representation ability by encoding knowledge from an unstructured domain-specific corpus. Zeng et al. \cite{zeng2022KV-PLM} proposed to encode both SMILES \cite{weininger1988smiles, weininger1989smiles} representation and text corpus of molecules simultaneously using a BERT \cite{devlin2018BERT} model as backbone. To train the proposed model, they extracted a corpus from the semantic scholar \cite{lo2019s2orc}, a database of over 136M published scientific literature, to construct 10k molecule-text pairs. Su et al. \cite{su2022MoMu} proposed to apply the contrastive learning paradigm with distinguishing graph and text branches. To overcome limited training data, authors utilized pre-trained models in the biochemical molecular domain for both GNN, GraphCL \cite{you2020GraphCL}, and text encoder, SciBERT \cite{beltagy2019scibert}. Liu et al. \cite{liu2023MoleculeSTM} pre-trained their model on a larger training set of 280k pairs, and both previous studies enabled zero-shot tasks and showed improved performance compared to unimodal models on downstream tasks.

\paragraph{Multimodal learning for crystals}
Multimodal learning in crystalline materials has not been explored extensively. Das et al. \cite{das2023crysmmnet} proposed CrysMMNet, which simultaneously integrates graph inputs and text inputs describing the structure of materials generated by a Robocrystallographer \cite{ganose2019robocrystallographer}. However, this approach focuses on creating joint representations and using them only to predict material properties and does not address issues related to zero-shot tasks and downstream tasks using natural language.

\section{Methods}
\subsection{Data collection}\label{sec:data_collection}
For training a language-materials multimodal contrastive learning model, we follow two steps: data preparation and model training. First, we use language-material pair data, which is language model-based synthetic data that Park et al. \cite{park2023materialsnarratives} have worked on previously. We also retrieve domain-specific natural language text from large-scale unlabeled materials in the SCI paper dataset. Afterward, the data of each modality passes through a graph encoder and a text encoder and is trained to be mapped to the joint representation space. In this way, our model can link materials graphs with materials engineering text descriptions. 

We collect 126k materials data from Materials Project\cite{jain2013MaterialsProject} to train our multimodal foundation model. To obtain crystal structure graphs of the collected compounds, we use the function to convert to crystal graphs from the Jarvis-tools library\cite{choudhary2020JARVIS}. To obtain the text for each material, we use 126k material-synthesized text pairs from our previous work\cite{park2023materialsnarratives}. We also search the corpus with weak supervision using the full text of 4M academic papers exclusively available to our group using chemical formulas as queries. In the collected data, there are a total of 400k material-natural language sentence pairs corresponding to approximately 400 inorganic compounds.

\subsubsection{Data augmentation}
Each crystal graph $\mathcal{G}=(V,E)$ undergo augmentation into augmented graph $\hat{\mathcal{G}}=(\hat{V}, \hat{E})$. In the pursuit of advancing the representative capacity of our model for crystal graphs, we have employed an augmentation strategy tailored for contrastive learning. To elucidate, let \( G = (V, E) \) represent the original crystal graph, where \( V \) and \( E \) denote the vertices and edges, respectively. The probabilistic augmentation process, denoted by \( \mathcal{A} \), involves perturbations of \( G \) to generate an augmented graph \( G' = \mathcal{A}(G,\xi') = (V', E') \), where $\xi$ denotes a random seed. These perturbations are designed to maintain the intrinsic geometric and chemical properties of the crystal structure, ensuring that \( G' \) remains within the manifold of physically plausible crystal graphs. We define \( \mathcal{A} \) such that it encompasses operations like edge removal \( \mathcal{A}_{er} \), node dropping \( \mathcal{A}_{nd} \), and subgraph sampling \( \mathcal{A}_{ss} \). The contrastive learning framework leverages a pair of augmented instances \( (G', G'') \) derived from the same original graph \( G \) to optimize the representation space, where \( G'' = \mathcal{A}(G,\xi'') \). This approach encourages the model to learn robust representations by enforcing invariance to the augmentations and, consequently, capturing the underlying topology and features critical to crystal graph representation.

In the domain of natural language processing, the efficacy of contrastive learning is significantly bolstered by the strategic augmentation of text data. For this purpose, we have implemented a token masking strategy, often denoted as \( \mathcal{T} \), to modify the textual input in a manner that is conducive to learning robust and discriminative features. Let a text sequence be represented by \( S = \{t_1, t_2, ..., t_n\} \), where \( t_i \) corresponds to the \( i^{th} \) token in the sequence. The probabilistic augmentation \( \mathcal{T}(S, \zeta') \) yields a new sequence \( S' = \{t'_1, t'_2, ..., t'_n\} \), wherein each token \( t'_i \) is either retained as the original token \( t_i \) with probability \( p \), or replaced with a special [MASK] token with probability \( 1-p \), signifying the application of token masking. ,$\zeta$ denotes a random seed. This process can be formally expressed as:

\begin{equation}
t'_i =
  \begin{cases} 
   t_i, & \text{with probability } p \\
   \text{[MASK]}, & \text{with probability } 1-p
  \end{cases}
\end{equation}

The contrastive learning paradigm leverages these augmented sequences \( S' \) and \( S'' \), where \( S'' = \mathcal{T}(S, \zeta'') \), both originating from the same initial sequence \( S \). Through this methodology, the model is trained to anchor the semantic representation of the text despite the perturbations introduced by token masking, thereby fostering an representation space that is invariant to such alterations and better captures the essence of the textual information.

\subsection{Graph encoder}
In the landscape of predictive analytics for material properties, Graph Neural Networks (GNNs) emerge as one of the most prominent methodologies, showcasing significant success \cite{xie2018CGCNN, choudhary2021ALIGNN}. Owing to their ability to encapsulate local environments and model atomic interactions, GNNs have propelled the field of material property prediction to new heights. A notable example is the Crystal Graph Convolutional Neural Network (CGCNN) \cite{xie2018CGCNN}, which enhances the predictive tasks for material properties by applying graph convolution operations to crystal graphs, thereby substantially outperforming conventional machine learning approaches. Recent advancements include the incorporation of line graphs to capture three-body or four-body interactions using angular information \cite{choudhary2021ALIGNN} and leveraging vector operations \cite{schutt2021painn} for enhanced representational learning. In our exploration of the influence of graph hidden representations on the CLaC model architecture, we employed CGCNN and PaiNN as graph encoders, facilitating a comparative study of their efficacy. Graph encoder can be formulated as follows:

\begin{equation}
    h_{\mathcal{G}}=f_g(\mathcal{G}, \left \{x_v|v\in V \right \}, \left \{ e_{uv}| (u,v)\in E\right \})
\end{equation}

where $\mathcal{G}=(V,E)$ is a graph with a set of nodes $V$ and a set of edges $E$, $x_v$ represents the feature vector of node $v$, $e_{uv}$ represents the feature vector of edge $(u,v)$, $f_g$ represents the GNN, and $h_{\mathcal{G}}$ represents hidden representation of graph after pooling. CGCNN embedding\cite{xie2018CGCNN} was selected as an initial node feature vector and radial basis function based on node displacement vector was selected as an initial edge feature.

\subsection{Text encoder}
The BERT structure has been widely used to extract meaningful features from text. Typically, it has been used to encode text in vision-language multimodal models such as DeCLIP and CLIP-lite\cite{li2021DeCLIP, shrivastava2023clip-lite}. Beltagy et al. \cite{beltagy2019scibert} presented SciBERT, which was pre-trained using 1.14M scientific papers in the biomedicine and computer science fields. Furthermore, Gupta et al.\cite{gupta2022matscibert} applied domain adaptive pre-training to SciBERT to develop a materials recognition language model. MatSciBERT trained on the materials science corpus with an additional vocabulary provides improved performance in several downstream tasks such as named entity recognition and classification. In this study, SciBERT and MatSciBERT are adopted as text encoders for comparative study. To prevent catastrophic forgetting of BERT-based models, we also add the masked language model loss term. Text encoders can be formulated as follows:

\begin{equation}
    h_{t}=f_t(\left \{x_i|i\in  \left \{ 1,...,n\right \}\right \})
\end{equation}

where $\left \{x_i|i\in  \left \{ 1,...,n\right \}\right \}$ represents the sequence of input token embeddings, $h_{t}$ is the sequence of hidden states produced by the BERT encoder, and $f_t$ denotes the BERT model.

\subsection{Optimization}
In this section, we describe our pre-training framework (Fig. \ref{fig:architecture}) for crystal graphic representation learning. Given a dataset of graph-text pairs, our goal is to train a graph encoder and a text encoder such that representations learned from each encoder share maximum mutual information. Consider a graph encoder network, \(f_{g}\) with parameter \(\theta_g\) and a text encoder \(f_{t}\) with parameter \(\theta_t\). Let \((x_g, x_t)\) is a sampled graph-text pair and \(f_{g}(x_g)\) and \(f_{t}(x_t)\) denote the representations extracted from the networks. 

For given random entities of crystal graph \(x_g\) and text \(x_t\), their mutual information is defined as a Kullback-Leibler (KL) divergence between their joint distribution \(p(x_g,x_t)\) and the product of their marginal distributions, \(p(x_g)p(x_t)\) as,

\begin{equation}
    I(x_g;x_t)=D_{\text{KL}}(p(x_g,x_t)\,||\,p(x_g)p(x_t)),
\end{equation}

For high-dimensional continuous variables, mutual information is notoriously difficult to estimate. We train it to maximize it using the Jensen-Shannon Divergence bound, similar to the formula used in generative modeling. This approach is differentiable and does not depend on the number of negative samples \cite{shrivastava2023clip-lite}.

\begin{equation}
I(h_{\mathcal{G}};h_{t})\geq\hat{I}_{\omega}^{JSD}(h_{\mathcal{G}};h_{t}):=\mathbb{E}_{P(h_{\mathcal{G}},h_{t})}[-\mathrm{log}(1+e^{-T_{\omega}})]-\mathbb{E}_{P(h_{\mathcal{G}})P(h_{t})}[\mathrm{log}(1+e^{T_{\omega}})],
\end{equation}

where \(T_\omega: \mathcal{X}_g \times \mathcal{X}_t \rightarrow \mathbb{R}\) is the discriminator neural network with trainable parameters \(\omega\) which are jointly optimized to distinguish between a paired-sample from a joint distribution (positive pair) and another pair from the product of marginals (negative pair). Therefore, our contrastive objective can be described as follows:

\begin{equation}
(\hat{\omega},\hat{\theta_{g}},\hat{\theta_{t}})=\underset{\omega,\theta_{g},\theta_{t}}{\mathrm{argmax}}\;\hat{I}^{JSD}_{\omega}(f_{g}(x_{g}),f_{t}(x_{t})),
\end{equation}

Finally, our loss function consisting of inter-modal and intra-modal alignment is the following equation.

\begin{equation*}
    \mathcal{L}_{inter-modal} = -\left( \mathbb{E}_{P(h_{\mathcal{G}}, h_{t})}\left[-\log(1 + e^{-T_{\omega}})\right] - \mathbb{E}_{P(h_{\mathcal{G}})P(h_{t})}\left[\log(1 + e^{T_{\omega}})\right] \right)
\end{equation*}
\begin{equation*}
    \mathcal{L}_{intra-modal, graph} =  -\left( \mathbb{E}_{P(h_{\mathcal{G}}, h_{\mathcal{G}'})}\left[-\log(1 + e^{-T_{\omega_g}})\right] - \mathbb{E}_{P(h_{\mathcal{G}})P(h_{\mathcal{G}'})}\left[\log(1 + e^{T_{\omega_g}})\right] \right)
\end{equation*}
\begin{equation*}
    \mathcal{L}_{intra-modal, text} =   -\left( \mathbb{E}_{P(h_{t}, h_{t'})}\left[-\log(1 + e^{-T_{\omega_t}})\right] - \mathbb{E}_{P(h_{t})P(h_{t'})}\left[\log(1 + e^{T_{\omega_t}})\right] \right)
\end{equation*}
\begin{equation*}
    \mathcal{L}_{MLM} = -\sum_{i \in \mathcal{M}} \log P(x_i \mid \mathbf{x}_{\backslash i})
\end{equation*}
\begin{equation}
    \mathcal{L}_{total} = \mathcal{L}_{inter-modal} + \mathcal{L}_{intra-modal, graph} + \mathcal{L}_{intra-modal, text} +\mathcal{L}_{MLM}
\end{equation}

where $\omega_g$ and $\omega_t$ represent parameters of discriminator neural network for graph and text intra-modality, respectively. $\mathcal{M}$ represents the set of masked positions in the input sequence. $x_i$ is the true token at the masked position $i$. $\mathbf{x}_{\backslash i}$ indicates the input sequence with the token at position $i$ masked out.

\section{Results}\label{results}
\subsection{Contrastive language-crystals pre-training model}  
\subsubsection{Overview} 
\begin{figure*}[h]
    \centering
    \includegraphics[width=\textwidth]{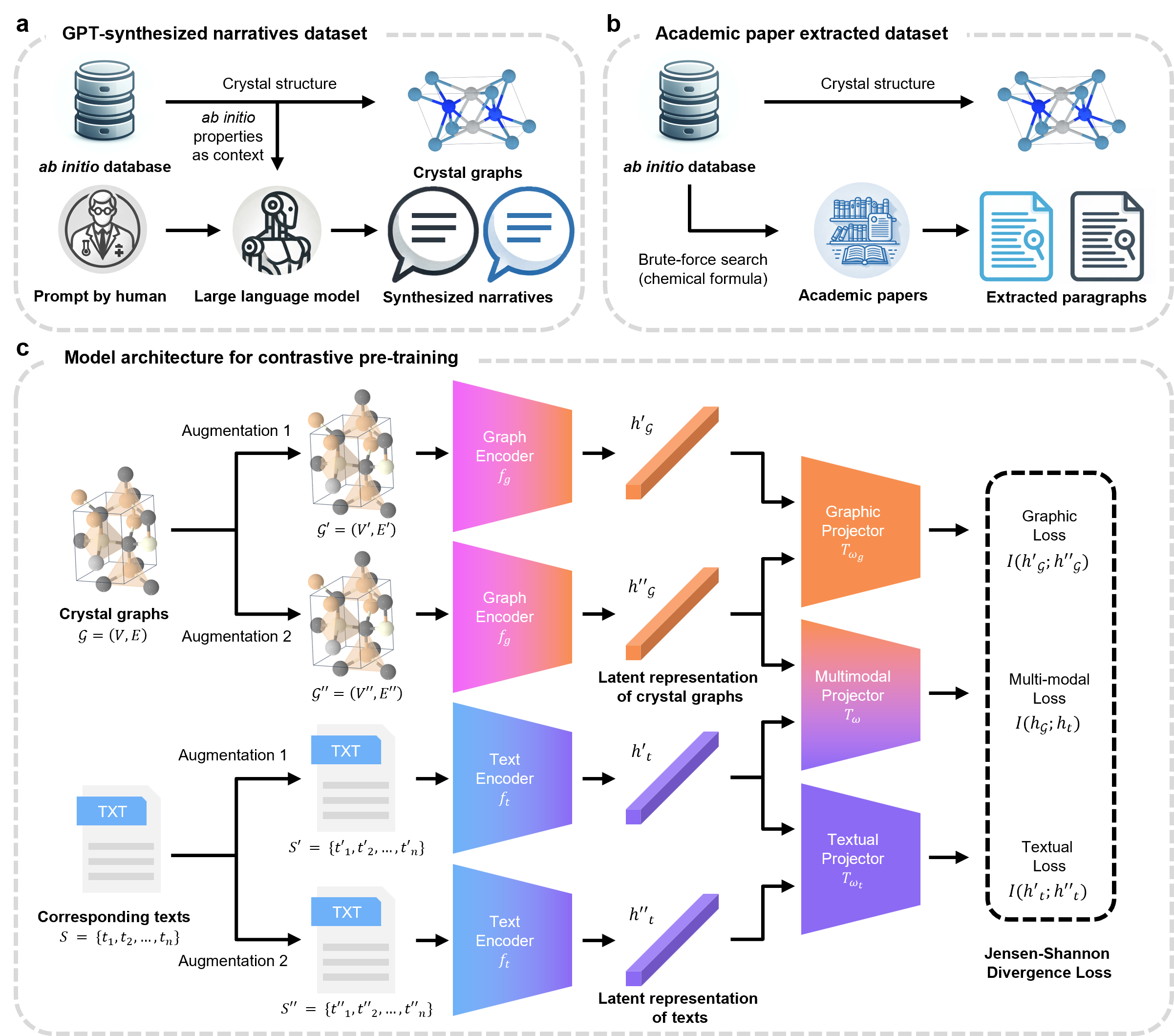}
    \caption{\textbf{Conceptual schematic diagram of the data pipeline and model architecture of the contrastive language-crystals multimodal pretraining.} \textbf{a}, The data processing flow for generating GPT-synthesized narrative multimodal pair data. \textbf{b}, The data processing flow for generating multimodal data from natural language academic papers. \textbf{c}, The overall architecture of the model. The original graph, text, and corresponding augmented entities pass through graph and text encoders, which feed each projector, leading to a multimodal projector that aligns both data types. }
    \label{fig:architecture}
\end{figure*}

In this section, we introduce the overall data pipeline, model architecture and the pre-training process as illustrated in Fig. \ref{fig:architecture}. We designed a foundation model for a comprehensive understanding of both natural language and crystal structures. The model consists of two neural network branches, a \textit{crystal graph encoder} and a \textit{text encoder}, which encodes the graphs and texts into a joint representation space, respectively. 

\subsubsection{Pre-training strategy} 
To pre-train our model for a multimodal alignment, we use GPT-synthesized narratives\cite{park2023materialsnarratives} for the Materials Project as well as academic paper paragraphs. These narratives are generated based on publicly available materials databases, leveraging the reasoning power of LLMs (Fig. \ref{fig:architecture}a). As an example, we present a generated narrative for a KNiIO$_6$ crystal (Supplementary Fig. \ref{fig:example_narrative}). We also perform a brute-force search to extract paragraphs containing specific chemical formulas from the academic literature (Fig. \ref{fig:architecture}b). GPT narratives contain 300 times more diverse crystals than can be found in the academic literature (details in Section \ref{sec:data_collection}) written by humans. This mitigates biases present in existing academic literature, as most researchers chase after a small number of so-called ``hot'' materials and neglect other possibilities on the thermodynamic convex hull, and allows the model to easily generalize its knowledge to a variety of downstream tasks. The dataset is split into 8:1:1 ratios for training, validation, and testing, respectively. 

We use inter-modal and intra-modal contrastive learning in this study by following a data-efficient multimodal contrastive pre-training approach \cite{shrivastava2023clip-lite} to maximize the mutual information (Fig. \ref{fig:architecture}c). Due to the scarcity of crystal-to-text data compared to image-to-text data, training both crystal-graph and text encoders from scratch is insufficient. To address this issue, Su et al.\cite{su2022MoMu} used the weights of the GraphCL \cite{you2020GraphCL}, pre-trained on biomedical molecules, and SciBERT \cite{beltagy2019scibert}, pre-trained on biomolecular corpora, as the initial weights of the graph and text encoder, respectively. However, unlike the biochemical molecular domain, where pre-trained GNNs trained on over 10 million unlabelled data\cite{fang2022GEM, zeng2022ImageMol, wang2022MolCLR}, research on crystal graphs \cite{koker2022CrystalCLR, magar2022CrystalTwins} have relied on only 126k Materials Project data. Therefore, we performed multimodal joint pre-training for both encoders on the prepared text-graph dataset: the graph encoder was trained from scratch, while the text encoder was initialized from a pre-trained model. 

\subsubsection{Evaluation}  
\begin{figure*}[h]
    \centering
    \includegraphics[width=\textwidth]{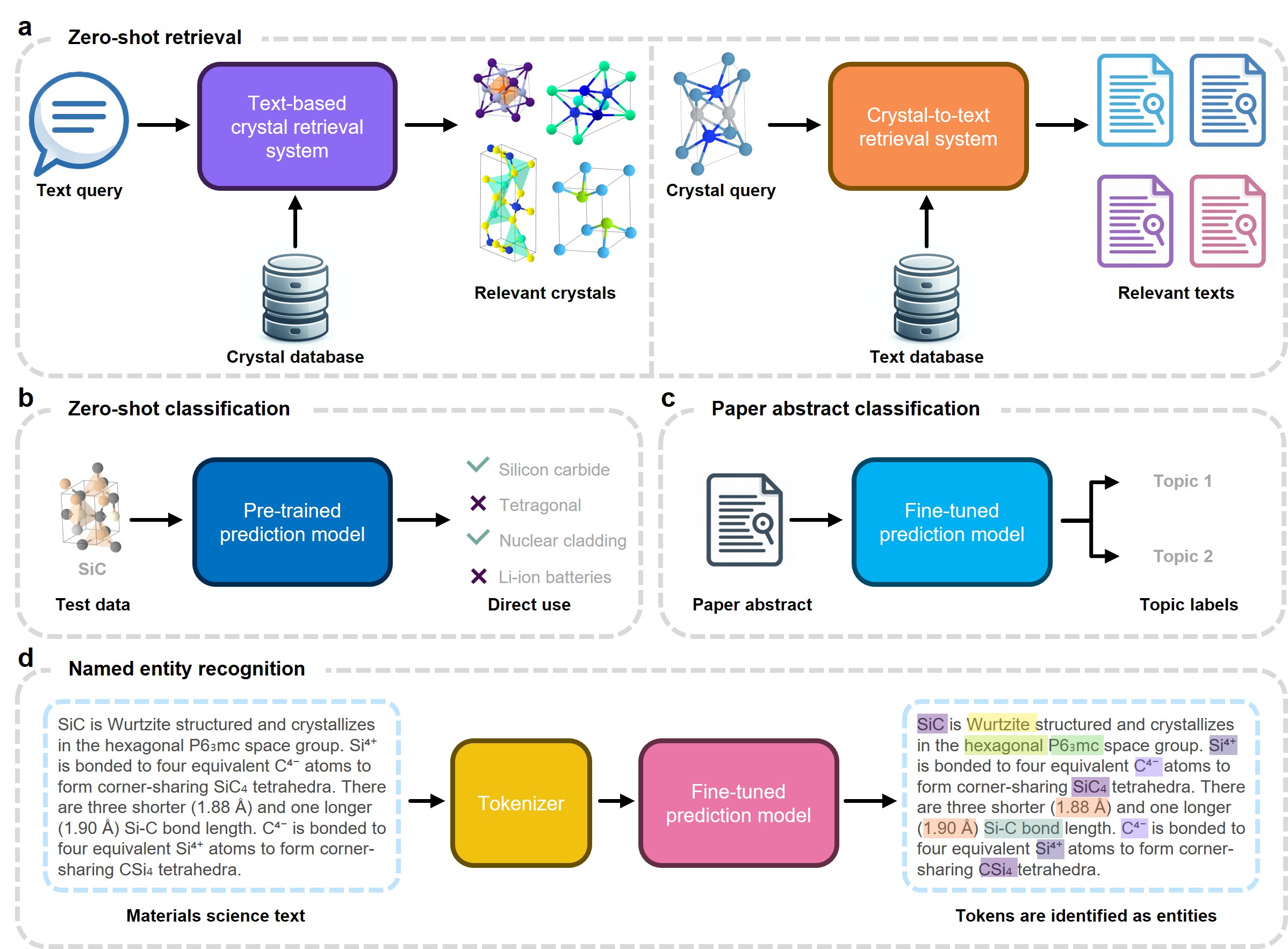}
    \caption{\textbf{Pipeline of downstream tasks.} \textbf{a}, Text-crystal zero-shot retrieval and crystal-text zero-shot retrieval. \textbf{b}, Zero-shot classification. \textbf{c}, Paper abstract classification. \textbf{d}, Named entity recognition. }    
    \label{fig:tasks}
\end{figure*}

Our pre-trained multimodal model can provide several utilities for crystal discovery. First, our model processes crystal graphs and text tokens simultaneously, enabling flexible information retrieval even for complex natural language queries. Therefore, it is not limited to pre-defined labels of crystals. Second, our pre-training strategy using inter-modal alignment helps large language models to achieve higher levels of comprehensive understanding regarding materials science, based on the 3D structural information. To emphasize these aspects, we propose to evaluate the model using zero-shot tasks and several natural language processing downstream tasks (Fig. \ref{fig:tasks}). 

For comparison, we set up several baseline models for each task. We also tested several graph encoders and text encoders to compare their effects on our model. A CGCNN\cite{xie2018CGCNN} and a polarizable atom interaction neural network (PaiNN)\cite{schutt2021painn} were considered the graph encoder, while a SciBERT\cite{beltagy2019scibert} and a MatSciBERT\cite{gupta2022matscibert} as the text encoder. CGCNN is one of the most widely used GNNs in crystalline materials, and PaiNN can improve performance with a small number of parameters by introducing equivariant representation.
\subsection{Zero-shot retrieval}  
Zero-shot retrieval refers to the capability of a model to retrieve relevant information from a database without having seen any examples of specific queries during its training phase. Pre-trained to maximize mutual information between crystal graphs and their corresponding descriptions, our model can identify relevant entities by matching the closest latent representations across different modalities. Graph-to-text retrieval aims to retrieve the most relevant text description for a given crystal graph. The text-to-graph task evaluates the model's ability to find a crystal from a database that best fits a given description. As a test dataset for the zero-shot retrieval, we use a test split of GPT-synthesized narratives dataset consisting of 12k crystal-text pairs. Since the test split of academic papers dataset often describes more than 2 materials in one paragraph, it is not proper to use it as a benchmark. The accuracy of retrieval is used as the metric.

We convert each material into a graph based on neighborhood-based graph construction. The test dataset is randomly sampled to consist of minibatches of size 1,024, meaning the candidate pool for each entity is 1,024. This is a remarkable extension, considering our model is trained on a batch size of 8. As the size of the candidate pool increases, the probability of the model correctly identifying the exact entity decreases, making the task considerably more challenging. Additionally, when the pool size increases to tens of thousands, it becomes challenging to fit this into the context length of an LLM. We calculate the cosine similarities between the representation of the query entity and all the other opposite modality entities in one batch. By ranking the cosine similarity, we can evaluate whether our model is able to retrieve their original pair (Supplementary Fig. \ref{fig:retrieval_explanation}). 

We proposed three approaches as baseline models (Supplementary Fig. \ref{fig:baseline_models}). The detailed explanation and discussion of baseline models are described in Supplementary Section \ref{sec:baseline_models}

We investigated the accuracy of various models in retrieving the crystal structures from given natural language descriptions, as summarized in Fig. \ref{fig:zero-shot}a. The model is considered accurate if it ranks the original crystal-text pair within the top $n$ when listing candidates in order of similarity from a pool of 1,024 crystals for a given text query. Letters after the CLaC model indicate the encoder configurations: `C' for CGCNN and `P' for PaiNN in the crystal graph domain, `S' for SciBERT and `M' for MatSciBERT in the textual domain. Our CLaC models demonstrate significant improvements in retrieval accuracy. Despite the substantial extension of batch size, our model can generalize its ability for multimodal retrievals. This suggests a substantial advancement in the model's ability to comprehend and predict complex material properties from textual data.

The test results show that when using the PM encoder, search accuracy increases compared to when using the CS encoder. This phenomenon underscores the potency of the PaiNN encoder in deciphering the structural intricacies of materials from their graphical representations. PaiNN is designed to use a direction-based message function to model complex four-body interactions and can perform better than CGCNN, which explicitly models two-body interactions. MatSciBERT also has enhanced capabilities in handling the complex semantics of material science narratives compared to SciBERT. CLaC-PM model, utilizing the PaiNN and MatSciBERT duo, achieves the highest accuracy with 82.50\%, 95.91\%, and 99.37\% for Top 1, Top 3, and Top 10 retrieval metrics on GPT narratives, respectively.
    
\subsection{Zero-shot multimodal understanding}  
Zero-shot understanding refers to the ability of an AI model to solve a new problem without prior specific training for that problem. For example, our model was not explicitly trained for crystal system classification, but it is expected to be able to perform that task due to the comprehensive and deep pre-training.

To evaluate zero-shot understanding, we designed zero-shot classification problems on the test dataset from the Materials Project dataset as described above. Here we extracted several statements with 4 subjects: composition, structure, composition-structure, and oxide type, which together describe the essential solid-state chemical properties of the crystal. The zero-shot multimodal retrieval task can be interpreted as a choose-one-multiple-choice problem. For details about generating multiple choices, please refer to Supplementary Section \ref{generating_multiple_choice}. The model is required to retrieve the textual description for a given crystalline structure. The accuracy of the choose-one-from-multiple-choice problem is used as the evaluation metric. Three aforementioned baseline models are also evaluated to compare zero-shot understanding performances. 

\begin{figure*}[h]
    \centering
    \includegraphics[width=0.9\textwidth]{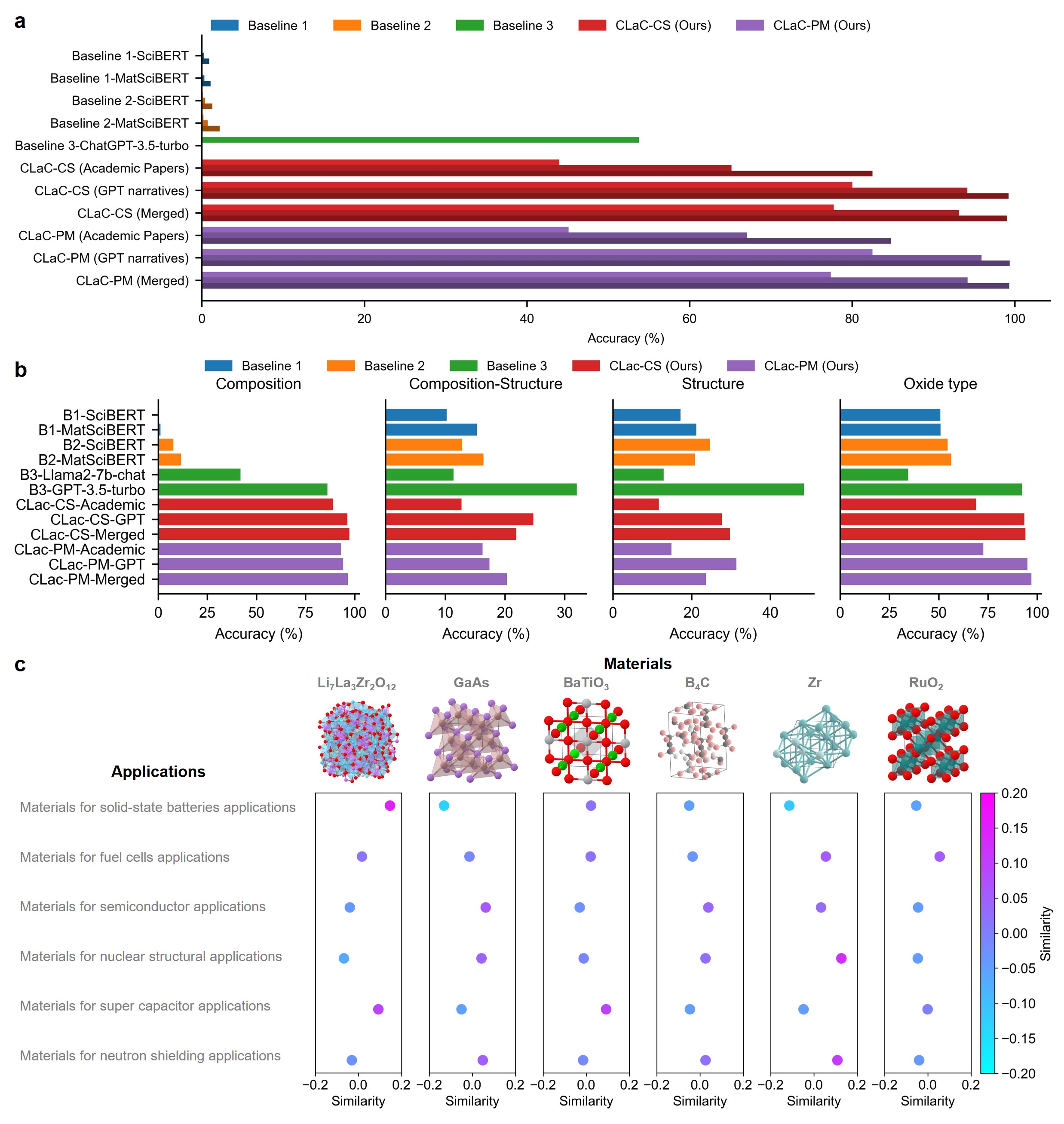}
    \caption{\textbf{Zero-shot generalization ability.} \textbf{a}, Text-to-crystal zero-shot retrieval accuracy within 1,024 candidates pool. `CS' signifies that the graph encoder is a CGCNN and the text encoder is a SciBERT. `P' indicates PaiNN, and `M' indicates MatSciBERT. \textbf{b}, Zero-shot classification accuracy. \textbf{c}, A graphical representation of the material-application similarity matrix. The matrix showcases six different materials – ${\rm Li_7La_3Zr_2O_{12}}$, ${\rm GaAs}$, ${\rm BaTiO_3}$, ${\rm B_4C}$, ${\rm Zr}$, and ${\rm RuO_2}$ – and their associations with six distinct application categories: solid-state batteries, fuel cells, semiconductors, nuclear structural materials, supercapacitors, and neutron shielding. Colored dots indicate the relevance of each material to the respective application, with the color coding corresponding to the degree of suitability based on underlying material properties.}
    \label{fig:zero-shot}
\end{figure*}

The zero-shot multimodal multiple-choice problem results are shown in Fig. \ref{fig:zero-shot}b. First, we observe that all our algorithms’ accuracies achieve remarkable levels of compositional understanding. We also observe that the `baseline 2' models perform as well as random guesses since the graph encoder is randomly initialized. On the other hand, the structural understanding of our model is not satisfactory. This may be caused by the difference between the degree of freedom of composition and structure. However, considering that the CLaC model has approximately 50 times fewer parameters compared to Llama2-7b, it is still a notable improvement. During contrastive learning, because of the small batch size, the model has more chance of exposure to negative samples consisting of different elements than negative samples with different crystal systems. Another possibility may be that relevant information is not sufficiently captured due to oversmoothing and oversquashing, which are typical shortcomings of GNN. Future work may consider using a Graph transformer to overcome oversquashing.

We illustrate our model's zero-shot performance through additional case studies. Instead of performing a classification task in the multiple-choice paradigm, we extend it to explore potential applications of materials. We can screen several materials suitable for specific applications by calculating similarities. The visual matrix in Fig. \ref{fig:zero-shot}c positions six materials along the vertical axis, each linked horizontally to a range of specialized application domains through a systematic array of colored dots. Each material is evaluated against six application categories: solid-state batteries, fuel cells, semiconductor devices, nuclear structural components, supercapacitors, and neutron shielding. The materials featured are ${\rm Li_7La_3Zr_2O_{12}}$, a lithium lanthanum zirconium oxide (LLZO) often utilized as a solid electrolyte in solid-state battery technology; Gallium arsenide (${\rm GaAs}$), a compound semiconductor renowned for its superior electron mobility compared to silicon; Barium titanate (${\rm BaTiO_3}$), widely acknowledged for its use in multilayer ceramic capacitors and electroceramics; Boron carbide (${\rm B_4C}$), one of the hardest materials available, serving as an ideal candidate for armor and neutron shielding due to its high hardness and low density; Zirconium (${\rm Zr}$), a lustrous metal that finds its applications in nuclear reactors owing to its low neutron-capture cross-section; and Ruthenium dioxide (${\rm RuO_2}$), a compound known for its applications in electrochemical capacitors, also known as supercapacitors, due to its excellent conductivity and durability. This zero-shot learning capability for specific industry sectors has potential applications in natural language-based target material generation. Our model represents the first step in this research direction. Realizing instruction-based material discovery using the developed model remains a challenge for future work.

\subsection{Named entity recognition}  
Named Entity Recognition (NER) is a subtask of information extraction that involves identifying and classifying named entities mentioned in text into predefined categories. Our multimodal pre-trained language models have the potential to be enhanced to better identify entities related to materials. We first evaluate CLaC on the NER task and present the experimental results of the materials science NER datasets, solid oxide fuel cells (SOFC)-slot\cite{walker2021SOFC-slot}, which have materials science domain-specific information. For comparison, we also conduct the same experiments for SciBERT and MatSciBERT to reproduce the results reported by Gupta et al.\cite{gupta2022matscibert}. 

\begin{figure*}[h]
    \centering
    \includegraphics[width=\textwidth]{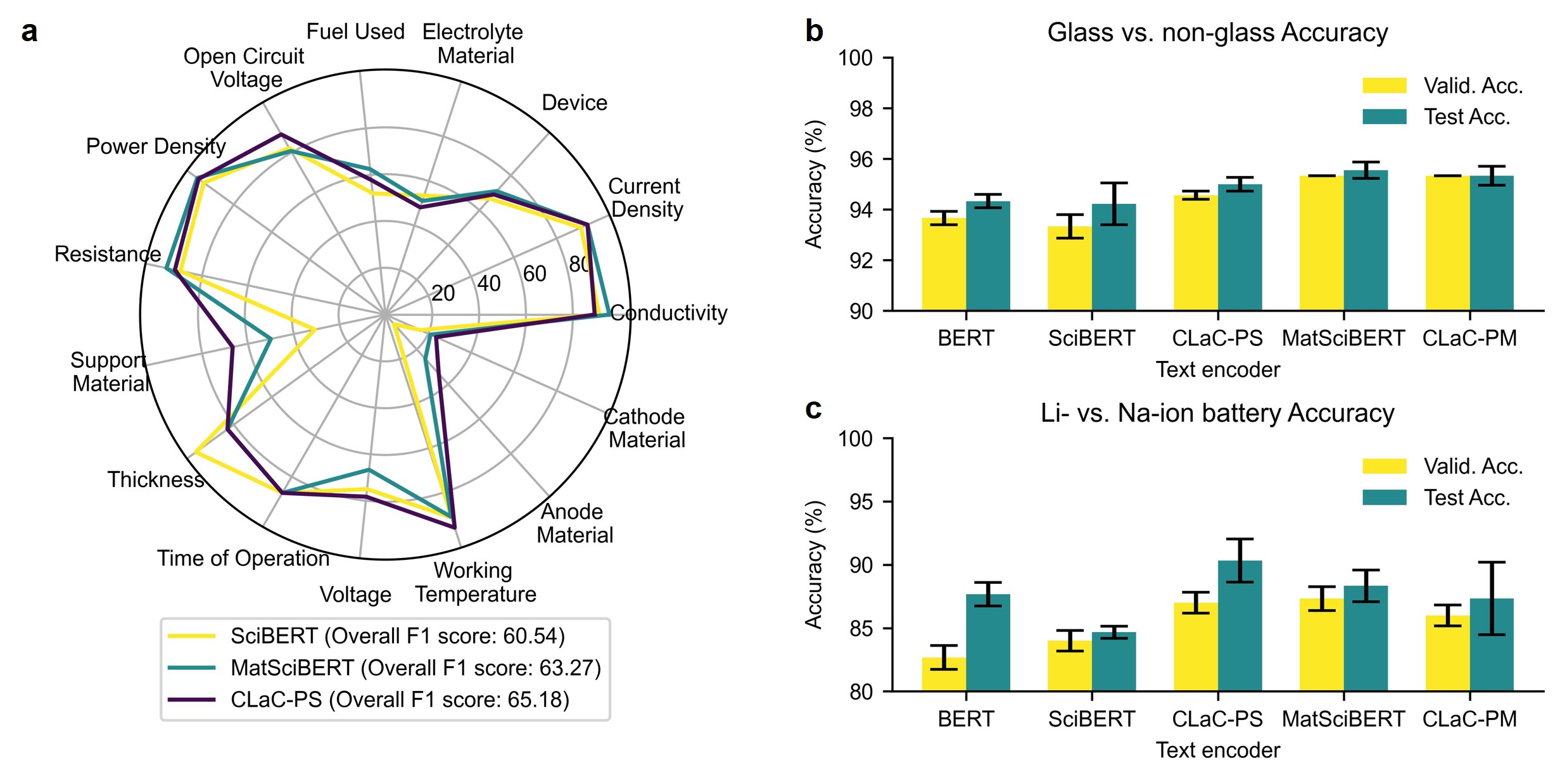}
    \caption{\textbf{Comparison of fine-tuning performance for downstream tasks.} \textbf{a}, Radar chart for named entity recognition models. The chart illustrates the evaluation metrics across various categories for three models: SciBERT, MatSciBERT, and CLaC with PaiNN and SciBERT encoders denoted as CLaC-PS. Each axis represents a different metric, and the distance from the center indicates the score achieved. The overall F1 scores for each model are highlighted. \textbf{b}, Validation accuracy (valid. acc.) and test accuracy (test acc.) of paper abstract classification performances for glass vs. non-glass and Li- vs. Na-ion battery. Each experiment averaged over three random seeds.}
    \label{fig:fine-tuning}
\end{figure*}

The radar chart in Fig. \ref{fig:fine-tuning}a provides a multi-dimensional comparison of performance metrics for three models: SciBERT, MatSciBERT, and CLaC-PS. The axes extending from the chart’s center describe specific entity types. The performance of each model is traced by the colored lines presenting a visual representation of the strengths and weaknesses of each model. The CLaC-PS model exhibits the highest average F1 score of 65.18, as indicated by the outermost boundary of its trajectory. This demonstrates that the joint-training strategy of our model improves the NER performance, which means the capability for accurately identifying and categorizing entities within scientific texts, than that of its base model. The F1 score for support material, anode material, and cathode material is especially superior to other text-only trained scientific NLP models. This suggests that language models trained on multimodal supervision of encoded crystal structures can acquire additional information that helps identify complex semantics in the materials science domain.

\subsection{Paper abstract classification} 
Paper abstract classification (PAC) by topic is a challenging task since it requires not only capturing the appearance of material-related vocabulary but also understanding complex context. In the PAC task, we evaluate the binary classification ability of a language model to classify paper abstracts. Here we use BERT, SciBERT, and MatSciBERT as baseline models. We used a publicly available dataset of paper abstracts that were annotated into two categories: glass and non-glass \cite{venugopal2021looking}. We also used our in-house annotated paper abstract dataset for the classification of Li- and Na-ion batteries topic. Classifying this topic is challenging since the model has to capture subtle differences among similar vocabularies. 

The validation and test accuracies, averaged over three random seeds, exhibit incremental performance enhancements, with CLaC-PS and CLaC-PM models achieving the first and second best test accuracies respectively, in the glass vs. non-glass category, and CLaC-PS outperforming all other models in the Li- vs. Na-ion battery classification as outlined in Fig. \ref{fig:fine-tuning}b. The evaluation results underscore the prowess of CLaC-PM, which not only surpasses the performance of the standard SciBERT model but also attains comparable results with MatSciBERT, achieving a 95.33\% accuracy on the test set for the glass vs. non-glass classification. For Li- vs. Na-ion battery topic classification, the CLaC-PS model demonstrates superior accuracy, indicating its capability to distinguish closely related terminologies regarding chemical elements. This empirical evidence presents the significant impact of incorporating multimodal architectures in refining the precision of language models for domain-specific tasks.

\subsection{Visualization of model attention} 

\begin{figure*}[h]
    \centering
    \includegraphics[width=\textwidth]{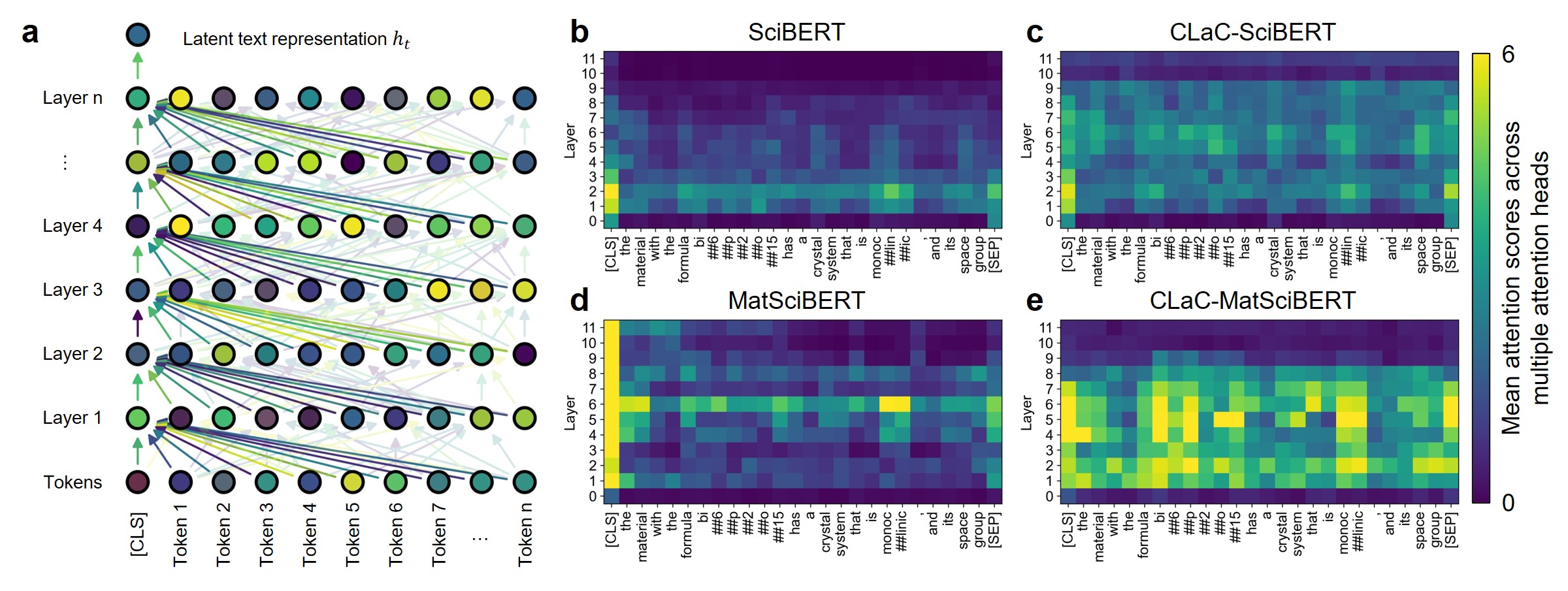}
    \caption{\textbf{Visualization of attention heatmaps from various models.} \textbf{a}, Schematic diagram of attention heatmap calculation. The four heatmaps represent attention patterns across different layers and heads of \textbf{b} SciBERT, \textbf{c} SciBERT as the text encoder of CLaC after joint training, \textbf{d} MatSciBERT, and \textbf{e} MatSciBERT as the text encoder of CLaC after joint training. Brighter colors indicate higher attention values, showing how the model focuses on different parts of the input sequence.}
    \label{fig:attn_beforeafter}
\end{figure*}

Overall, we show that CLaC, jointly trained on multimodal entities, can perform better than SciBERT on natural language processing-based downstream tasks. It is particularly notable that these training strategies are comparable to the expected performance when trained with novel tokens from the scientific literature in the materials domain, such as MatSciBERT. To understand this improvement, we investigated attention distribution across layers, which provides insights into how a model processes information hierarchically, before and after undergoing joint multimodal training (Fig. \ref{fig:attn_beforeafter}). For visualization, we calculate the attention for the CLS token across all heads (Fig. \ref{fig:attn_beforeafter}a). The attention values are summed over all the heads. By summing across the heads, we reduce complexity and can visualize which tokens the model is focusing on overall, without needing to consider individual attention heads. 

Before the multimodal joint training using CLaC (Fig. \ref{fig:attn_beforeafter} b and d), a comparison between SciBERT and MatSciBERT reveals distinct patterns in their attention mechanisms. Specifically, MatSciBERT displays activation of attention at higher layers for the CLS token in response to the same sentences, indicating its better pattern recognition capacity for materials science literature. In contrast, SciBERT, which has not been fine-tuned on materials science corpora, shows less pronounced attention in these higher layers. Upon applying CLaC for multimodal joint training, SciBERT begins to exhibit activation patterns in its 6th layer that are similar to those of MatSciBERT (Fig. \ref{fig:attn_beforeafter} c and e). This adaptation suggests that SciBERT has acquired a more nuanced understanding of the domain-specific content, enabling it to perform competitively on various NLP downstream tasks previously dominated by MatSciBERT. When the same CLaC training is extended to MatSciBERT, the enhancements are not as dramatic as those observed with SciBERT. This may be because MatSciBERT has already achieved a high level of baseline performance in tasks related to materials science.

\subsection{Visualization of latent space} 
\begin{figure*}[h]
    \centering
    \includegraphics[width=\textwidth]{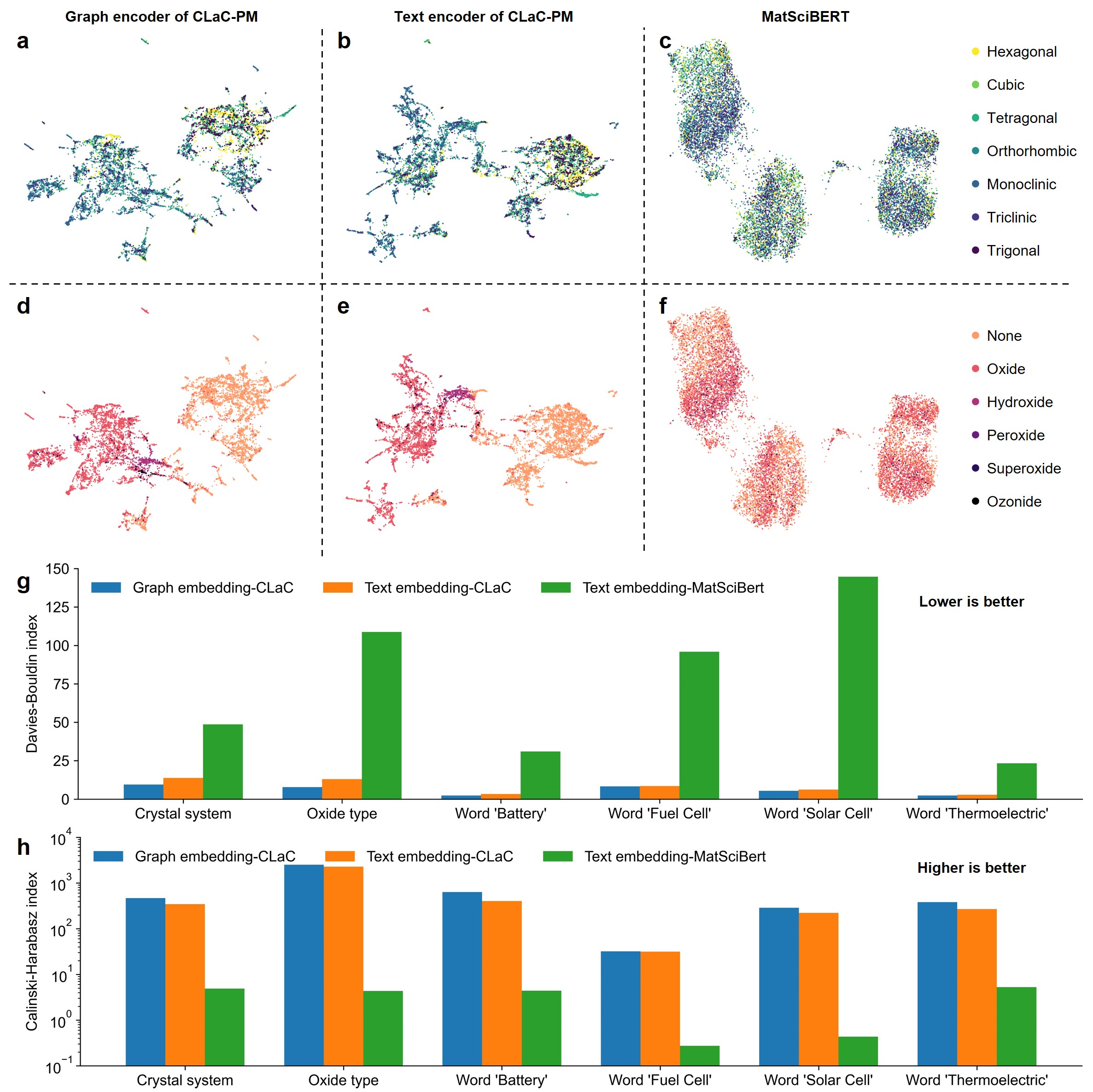}
    \caption{\textbf{Latent embeddings of the CLAC.} UMAP visualization of graph and text representations. Each column corresponds to graph representation (\textbf{a, d}), text representation after multimodal training (\textbf{b, e}), and text representation of pristine MatSciBERT (\textbf{c, f}). Each row corresponds to the crystal system classification (\textbf{a, b, c}) and the oxide type classification (\textbf{d, e, f}). \textbf{g} Comparison of Davies-Bouldin index for different labels. \textbf{h} Comparison of Calinski-Harabasz index for different labels.}
    \label{fig:embedding}
\end{figure*}

We visualized hi-dimensional representations of crystal graph and text entities of GPT-synthesized narratives using UMAP\cite{mcinnes2018umap} in Fig. \ref{fig:embedding}. UMAP is a dimensionality reduction technique that is particularly adept at preserving both local and global structures, and is suitable for visualizing high-dimensional data in two dimensions. In the presented UMAP visualization, we can show the successful clustering of materials by graph encoder and text encoder after multimodal training. The separation of clusters suggests that materials can be categorized based on their graph-based features, which might include atomic configurations, crystal structures, or other material properties represented in the graph data. The visualized representations are categorized for distinguishing oxide types as well as crystal systems. They also cluster a range of materials, including those with application-specific terminologies such as `battery', `fuel cell', `solar cell' or `thermoelectric' within the text (Supplementary Figure \ref{fig:embedding_battery}, \ref{fig:embedding_fuelcell}, \ref{fig:embedding_solarcell}, and \ref{fig:embedding_thermoelectric}). These clusterings have the potential to help find new materials suitable for specific applications without having to rely on calculating specific material properties. The plot illustrates that both graph and text representations are capable of discerning these categories, showcasing the effectiveness of the models in grouping similar crystals in the latent space. In contrast, before multimodal training, MatSciBERT did not exhibit the same level of distinction between these categories. This suggests that leveraging both graph and text data could provide a comprehensive understanding of material properties and their relationships, facilitating better material discovery and characterization.

We aimed to quantitatively evaluate whether various materials are well-distinguished within these latent embedding spaces. To achieve this, we calculated scores for different labels using the Davies-Bouldin (DB) index and the Calinski-Harabasz (CH) index. A lower DB score generally indicates better clustering performance, as it reflects more compact and well-separated clusters. Consistently achieving lower DB scores across diverse topics suggests that the embeddings effectively capture the underlying data structure. The CH index, on the other hand, measures the ratio of inter-cluster variance to intra-cluster variance. As a variance-based metric, higher CH index values indicate better clustering quality. Additionally, the results from the two encoders exhibit quite similar patterns, which seem to contribute to their strong performance in tasks like zero-shot retrieval. Combined with previous evaluations on zero-shot tasks and downstream tasks, these results highlight the superior embedding quality of CLaC caused by joint contrastive learning.

The visualization results illustrate that the CLaC model effectively maps materials and text into the latent space. However, there remains significant potential for further improvement. Specifically, we conducted an analysis using the DB index, CH index, and Silhouette score. Despite the low DB index and high CH index indicating improved clustering performance, the Silhouette score was found to be very close to 0. This outcome is likely attributed to noise present in the current labels and the high dimensionality of the latent space. The training data was generated using the inference capabilities of GPT-3.5, which likely introduced a non-negligible amount of noise into the dataset. Future research could focus on enhancing data generation methodologies and refining training processes to achieve a more precise alignment between the linguistic hypothesis space of scientists and the materials space, thereby enabling novel approaches for exploring this intersection.

\section{Discussion}  
In conclusion, we developed a crystal graph-aware multimodal language model that is trained on crystal graphs and materials science corpora, including GPT-synthesized narratives.  Through a series of experiments on zero-shot tasks and downstream tasks using standard benchmarks, we confirmed that leveraging synthetic data is a promising approach to addressing data scarcity and bias within the materials science domain. The multimodal model, trained with 3D crystal structure information from synthetic data, enabled the execution of zero-shot tasks, which were challenging to achieve with previous language models. Attention visualization patterns also demonstrate that transformer-based models, such as SciBERT, can similarly benefit from multimodal learning approaches, as seen in text-based pre-training for the materials science domain. Our model can serve as a baseline for future work, for example, providing similarity metrics for evaluating text-based crystal generation tasks.

Despite the strengths exhibited by our current models, there are limitations; Our model is currently only applicable to crystalline materials. However, contemporary materials science research handles polycrystals or materials with defects that affect their properties, rather than single crystals. Furthermore, the model cannot handle metal-organic frameworks at this stage. Therefore, our next step is to cultivate models capable of identifying correlations that encompass a broader spectrum of materials and structures, thereby achieving a higher level of comprehension. Such advancements will require the development of representations that can simultaneously handle not only single crystals but also high-dimensional structures such as polycrystals and defects. Additionally, advancements in data synthesis methods are needed to minimize hallucinations and generate more informative data. Another challenge is the lack of a standardized nomenclature in the field of inorganic materials science, unlike in the molecular domain. For example, IUPAC names naturally included in academic literature provide opportunities to learn various concepts from the surrounding context. On the other hand, terms like SiO$_2$ in the literature cannot be distinguished by nomenclature alone as to whether they refer to a crystal or non-crystal, bulk material, or nanoparticles. Future research should carefully address these limitations.

\bibliographystyle{unsrtnat}
\bibliography{references}  

\begin{thebibliography}{53}
\providecommand{\natexlab}[1]{#1}
\providecommand{\url}[1]{\texttt{#1}}
\expandafter\ifx\csname urlstyle\endcsname\relax
  \providecommand{\doi}[1]{doi: #1}\else
  \providecommand{\doi}{doi: \begingroup \urlstyle{rm}\Url}\fi

\bibitem[Kim et~al.(2017)Kim, Huang, Jegelka, and Olivetti]{kim2017VirtualScreening}
Edward Kim, Kevin Huang, Stefanie Jegelka, and Elsa Olivetti.
\newblock Virtual screening of inorganic materials synthesis parameters with deep learning.
\newblock \emph{npj Computational Materials}, 3\penalty0 (1):\penalty0 53, 2017.

\bibitem[Sorkun et~al.(2020)Sorkun, Astruc, Koelman, and Er]{sorkun2020AI-VirtualScreening}
Murat~Cihan Sorkun, S{\'e}verin Astruc, JM~Vianney~A Koelman, and S{\"u}leyman Er.
\newblock An artificial intelligence-aided virtual screening recipe for two-dimensional materials discovery.
\newblock \emph{npj Computational Materials}, 6\penalty0 (1):\penalty0 106, 2020.

\bibitem[Xie and Grossman(2018)]{xie2018CGCNN}
Tian Xie and Jeffrey~C Grossman.
\newblock Crystal graph convolutional neural networks for an accurate and interpretable prediction of material properties.
\newblock \emph{Physical review letters}, 120\penalty0 (14):\penalty0 145301, 2018.

\bibitem[Choudhary and DeCost(2021)]{choudhary2021ALIGNN}
Kamal Choudhary and Brian DeCost.
\newblock Atomistic line graph neural network for improved materials property predictions.
\newblock \emph{npj Computational Materials}, 7\penalty0 (1):\penalty0 185, 2021.

\bibitem[Kaundinya et~al.(2022)Kaundinya, Choudhary, and Kalidindi]{kaundinya2022predictionALIGNN}
Prathik~R Kaundinya, Kamal Choudhary, and Surya~R Kalidindi.
\newblock Prediction of the electron density of states for crystalline compounds with atomistic line graph neural networks (alignn).
\newblock \emph{arXiv preprint arXiv:2201.08348}, 2022.

\bibitem[Hoffmann et~al.(2019)Hoffmann, Maestrati, Sawada, Tang, Sellier, and Bengio]{hoffmann2019Voxel-VAE}
Jordan Hoffmann, Louis Maestrati, Yoshihide Sawada, Jian Tang, Jean~Michel Sellier, and Yoshua Bengio.
\newblock Data-driven approach to encoding and decoding 3-d crystal structures.
\newblock \emph{arXiv preprint arXiv:1909.00949}, 2019.

\bibitem[Noh et~al.(2019)Noh, Kim, Stein, Sanchez-Lengeling, Gregoire, Aspuru-Guzik, and Jung]{noh2019inverseMatter}
Juhwan Noh, Jaehoon Kim, Helge~S Stein, Benjamin Sanchez-Lengeling, John~M Gregoire, Alan Aspuru-Guzik, and Yousung Jung.
\newblock Inverse design of solid-state materials via a continuous representation.
\newblock \emph{Matter}, 1\penalty0 (5):\penalty0 1370--1384, 2019.

\bibitem[Court et~al.(2020)Court, Yildirim, Jain, and Cole]{court2020Cond-DFC-VAE}
Callum~J Court, Batuhan Yildirim, Apoorv Jain, and Jacqueline~M Cole.
\newblock 3-d inorganic crystal structure generation and property prediction via representation learning.
\newblock \emph{Journal of Chemical Information and Modeling}, 60\penalty0 (10):\penalty0 4518--4535, 2020.

\bibitem[Xie et~al.(2021)Xie, Fu, Ganea, Barzilay, and Jaakkola]{xie2021CDVAE}
Tian Xie, Xiang Fu, Octavian-Eugen Ganea, Regina Barzilay, and Tommi Jaakkola.
\newblock Crystal diffusion variational autoencoder for periodic material generation.
\newblock \emph{arXiv preprint arXiv:2110.06197}, 2021.

\bibitem[Lyngby and Thygesen(2022)]{lyngby2022CDVAE2Ddiscovery}
Peder Lyngby and Kristian~Sommer Thygesen.
\newblock Data-driven discovery of 2d materials by deep generative models.
\newblock \emph{npj Computational Materials}, 8\penalty0 (1):\penalty0 232, 2022.

\bibitem[Gong et~al.(2023)Gong, Yan, Xie, Shao-Horn, Gomez-Bombarelli, Ji, and Grossman]{gong2023GNNforcrystal}
Sheng Gong, Keqiang Yan, Tian Xie, Yang Shao-Horn, Rafael Gomez-Bombarelli, Shuiwang Ji, and Jeffrey~C Grossman.
\newblock Examining graph neural networks for crystal structures: limitations and opportunities for capturing periodicity.
\newblock \emph{Science Advances}, 9\penalty0 (45):\penalty0 eadi3245, 2023.

\bibitem[Pakornchote et~al.(2024)Pakornchote, Choomphon-Anomakhun, Arrerut, Atthapak, Khamkaeo, Chotibut, and Bovornratanaraks]{pakornchote2024dpcdvae}
Teerachote Pakornchote, Natthaphon Choomphon-Anomakhun, Sorrjit Arrerut, Chayanon Atthapak, Sakarn Khamkaeo, Thiparat Chotibut, and Thiti Bovornratanaraks.
\newblock Diffusion probabilistic models enhance variational autoencoder for crystal structure generative modeling.
\newblock \emph{Scientific Reports}, 14\penalty0 (1):\penalty0 1275, 2024.

\bibitem[Lookman et~al.(2019)Lookman, Balachandran, Xue, and Yuan]{lookman2019ALinMaterials}
Turab Lookman, Prasanna~V Balachandran, Dezhen Xue, and Ruihao Yuan.
\newblock Active learning in materials science with emphasis on adaptive sampling using uncertainties for targeted design.
\newblock \emph{npj Computational Materials}, 5\penalty0 (1):\penalty0 21, 2019.

\bibitem[Chen and Gu(2020)]{chen2020GenerativeInverseDesignUsingAL}
Chun-Teh Chen and Grace~X Gu.
\newblock Generative deep neural networks for inverse materials design using backpropagation and active learning.
\newblock \emph{Advanced Science}, 7\penalty0 (5):\penalty0 1902607, 2020.

\bibitem[Butler et~al.(2018)Butler, Davies, Cartwright, Isayev, and Walsh]{butler2018MLforMolecularandMaterialsScience}
Keith~T Butler, Daniel~W Davies, Hugh Cartwright, Olexandr Isayev, and Aron Walsh.
\newblock Machine learning for molecular and materials science.
\newblock \emph{Nature}, 559\penalty0 (7715):\penalty0 547--555, 2018.

\bibitem[Agrawal and Choudhary(2016)]{agrawal2016FourthParadigmPerspective}
Ankit Agrawal and Alok Choudhary.
\newblock Perspective: Materials informatics and big data: Realization of the “fourth paradigm” of science in materials science.
\newblock \emph{Apl Materials}, 4\penalty0 (5), 2016.

\bibitem[Devlin et~al.(2018)Devlin, Chang, Lee, and Toutanova]{devlin2018BERT}
Jacob Devlin, Ming-Wei Chang, Kenton Lee, and Kristina Toutanova.
\newblock Bert: Pre-training of deep bidirectional transformers for language understanding.
\newblock \emph{arXiv preprint arXiv:1810.04805}, 2018.

\bibitem[Radford et~al.(2019)Radford, Wu, Child, Luan, Amodei, Sutskever, et~al.]{radford2019GPT2}
Alec Radford, Jeffrey Wu, Rewon Child, David Luan, Dario Amodei, Ilya Sutskever, et~al.
\newblock Language models are unsupervised multitask learners.
\newblock \emph{OpenAI blog}, 1\penalty0 (8):\penalty0 9, 2019.

\bibitem[Kolesnikov et~al.(2019)Kolesnikov, Zhai, and Beyer]{kolesnikov2019SSLVision}
Alexander Kolesnikov, Xiaohua Zhai, and Lucas Beyer.
\newblock Revisiting self-supervised visual representation learning.
\newblock In Larry Davis, Philip Torr, and Song-Chun Zhu, editors, \emph{Proceedings of the IEEE/CVF conference on computer vision and pattern recognition}, pages 1920--1929, 2019.

\bibitem[Wang et~al.(2022)Wang, Wang, Cao, and Barati~Farimani]{wang2022MolCLR}
Yuyang Wang, Jianren Wang, Zhonglin Cao, and Amir Barati~Farimani.
\newblock Molecular contrastive learning of representations via graph neural networks.
\newblock \emph{Nature Machine Intelligence}, 4\penalty0 (3):\penalty0 279--287, 2022.

\bibitem[Fang et~al.(2022)Fang, Liu, Lei, He, Zhang, Zhou, Wang, Wu, and Wang]{fang2022GEM}
Xiaomin Fang, Lihang Liu, Jieqiong Lei, Donglong He, Shanzhuo Zhang, Jingbo Zhou, Fan Wang, Hua Wu, and Haifeng Wang.
\newblock Geometry-enhanced molecular representation learning for property prediction.
\newblock \emph{Nature Machine Intelligence}, 4\penalty0 (2):\penalty0 127--134, 2022.

\bibitem[Koker et~al.(2022)Koker, Quigley, Spaeth, Frey, and Li]{koker2022CrystalCLR}
Teddy Koker, Keegan Quigley, Will Spaeth, Nathan~C Frey, and Lin Li.
\newblock Graph contrastive learning for materials.
\newblock \emph{arXiv preprint arXiv:2211.13408}, 2022.

\bibitem[Magar et~al.(2022)Magar, Wang, and Barati~Farimani]{magar2022CrystalTwins}
Rishikesh Magar, Yuyang Wang, and Amir Barati~Farimani.
\newblock Crystal twins: self-supervised learning for crystalline material property prediction.
\newblock \emph{npj Computational Materials}, 8\penalty0 (1):\penalty0 231, 2022.

\bibitem[Radford et~al.(2021)Radford, Kim, Hallacy, Ramesh, Goh, Agarwal, Sastry, Askell, Mishkin, Clark, Krueger, and Sutskever]{radford2021CLIP}
Alec Radford, Jong~Wook Kim, Chris Hallacy, Aditya Ramesh, Gabriel Goh, Sandhini Agarwal, Girish Sastry, Amanda Askell, Pamela Mishkin, Jack Clark, Gretchen Krueger, and Ilya Sutskever.
\newblock Learning transferable visual models from natural language supervision.
\newblock In Marina Meila and Tong Zhang, editors, \emph{Proceedings of the 38th International Conference on Machine Learning}, volume 139 of \emph{Proceedings of Machine Learning Research}, pages 8748--8763. PMLR, 18--24 Jul 2021.
\newblock URL \url{https://proceedings.mlr.press/v139/radford21a.html}.

\bibitem[Rombach et~al.(2022)Rombach, Blattmann, Lorenz, Esser, and Ommer]{rombach2022StableDiffusion}
Robin Rombach, Andreas Blattmann, Dominik Lorenz, Patrick Esser, and Bj{\"o}rn Ommer.
\newblock High-resolution image synthesis with latent diffusion models.
\newblock In Rama Chellappa, Jiri Matas, Long Quan, and Mubarak Shah, editors, \emph{Proceedings of the IEEE/CVF conference on computer vision and pattern recognition}, pages 10684--10695, 2022.

\bibitem[Ramesh et~al.(2021)Ramesh, Pavlov, Goh, Gray, Voss, Radford, Chen, and Sutskever]{ramesh2021Dalle}
Aditya Ramesh, Mikhail Pavlov, Gabriel Goh, Scott Gray, Chelsea Voss, Alec Radford, Mark Chen, and Ilya Sutskever.
\newblock Zero-shot text-to-image generation.
\newblock In Marina Meila and Tong Zhang, editors, \emph{Proceedings of the 38th International Conference on Machine Learning}, volume 139 of \emph{Proceedings of Machine Learning Research}, pages 8821--8831. PMLR, 18--24 Jul 2021.
\newblock URL \url{https://proceedings.mlr.press/v139/ramesh21a.html}.

\bibitem[Liu et~al.(2023{\natexlab{a}})Liu, Li, Wu, and Lee]{liu2023llava}
Haotian Liu, Chunyuan Li, Qingyang Wu, and Yong~Jae Lee.
\newblock Visual instruction tuning.
\newblock \emph{arXiv preprint arXiv:2304.08485}, 2023{\natexlab{a}}.

\bibitem[Li et~al.(2023)Li, Wong, Zhang, Usuyama, Liu, Yang, Naumann, Poon, and Gao]{li2023llava-med}
Chunyuan Li, Cliff Wong, Sheng Zhang, Naoto Usuyama, Haotian Liu, Jianwei Yang, Tristan Naumann, Hoifung Poon, and Jianfeng Gao.
\newblock Llava-med: Training a large language-and-vision assistant for biomedicine in one day.
\newblock \emph{arXiv preprint arXiv:2306.00890}, 2023.

\bibitem[Irwin and Shoichet(2005)]{irwin2005zinc}
John~J Irwin and Brian~K Shoichet.
\newblock Zinc- a free database of commercially available compounds for virtual screening.
\newblock \emph{Journal of chemical information and modeling}, 45\penalty0 (1):\penalty0 177--182, 2005.

\bibitem[Belsky et~al.(2002)Belsky, Hellenbrandt, Karen, and Luksch]{belsky2002ICSD}
Alec Belsky, Mariette Hellenbrandt, Vicky~Lynn Karen, and Peter Luksch.
\newblock New developments in the inorganic crystal structure database (icsd): accessibility in support of materials research and design.
\newblock \emph{Acta Crystallographica Section B: Structural Science}, 58\penalty0 (3):\penalty0 364--369, 2002.

\bibitem[Jain et~al.(2013)Jain, Ong, Hautier, Chen, Richards, Dacek, Cholia, Gunter, Skinner, Ceder, et~al.]{jain2013MaterialsProject}
Anubhav Jain, Shyue~Ping Ong, Geoffroy Hautier, Wei Chen, William~Davidson Richards, Stephen Dacek, Shreyas Cholia, Dan Gunter, David Skinner, Gerbrand Ceder, et~al.
\newblock The materials project: A materials genome approach to accelerating materials innovation, apl mater.
\newblock \emph{APL Materials}, 2013.

\bibitem[Park et~al.(2023)Park, Jerng, Park, Kwon, Hsu, Ren, Yoon, and Li]{park2023materialsnarratives}
Yang~Jeong Park, Sung~Eun Jerng, Jin-Sung Park, Choah Kwon, Chia-Wei Hsu, Zhichu Ren, Sungroh Yoon, and Ju~Li.
\newblock 1.5 million materials narratives generated by chatbots.
\newblock \emph{arXiv preprint arXiv:2308.13687}, 2023.

\bibitem[Gilmer et~al.(2017)Gilmer, Schoenholz, Riley, Vinyals, and Dahl]{gilmer2017MPNN}
Justin Gilmer, Samuel~S. Schoenholz, Patrick~F. Riley, Oriol Vinyals, and George~E. Dahl.
\newblock Neural message passing for quantum chemistry.
\newblock In Doina Precup and Yee~Whye Teh, editors, \emph{Proceedings of the 34th International Conference on Machine Learning}, volume~70 of \emph{Proceedings of Machine Learning Research}, pages 1263--1272. PMLR, 06--11 Aug 2017.
\newblock URL \url{https://proceedings.mlr.press/v70/gilmer17a.html}.

\bibitem[Zbontar et~al.(2021)Zbontar, Jing, Misra, LeCun, and Deny]{zbontar2021barlowtwins}
Jure Zbontar, Li~Jing, Ishan Misra, Yann LeCun, and Stephane Deny.
\newblock Barlow twins: Self-supervised learning via redundancy reduction.
\newblock In Marina Meila and Tong Zhang, editors, \emph{Proceedings of the 38th International Conference on Machine Learning}, volume 139 of \emph{Proceedings of Machine Learning Research}, pages 12310--12320. PMLR, 18--24 Jul 2021.
\newblock URL \url{https://proceedings.mlr.press/v139/zbontar21a.html}.

\bibitem[Zeng et~al.(2022{\natexlab{a}})Zeng, Yao, Liu, and Sun]{zeng2022KV-PLM}
Zheni Zeng, Yuan Yao, Zhiyuan Liu, and Maosong Sun.
\newblock A deep-learning system bridging molecule structure and biomedical text with comprehension comparable to human professionals.
\newblock \emph{Nature communications}, 13\penalty0 (1):\penalty0 862, 2022{\natexlab{a}}.

\bibitem[Weininger(1988)]{weininger1988smiles}
David Weininger.
\newblock Smiles, a chemical language and information system. 1. introduction to methodology and encoding rules.
\newblock \emph{Journal of chemical information and computer sciences}, 28\penalty0 (1):\penalty0 31--36, 1988.

\bibitem[Weininger et~al.(1989)Weininger, Weininger, and Weininger]{weininger1989smiles}
David Weininger, Arthur Weininger, and Joseph~L Weininger.
\newblock Smiles. 2. algorithm for generation of unique smiles notation.
\newblock \emph{Journal of chemical information and computer sciences}, 29\penalty0 (2):\penalty0 97--101, 1989.

\bibitem[Lo et~al.(2019)Lo, Wang, Neumann, Kinney, and Weld]{lo2019s2orc}
Kyle Lo, Lucy~Lu Wang, Mark Neumann, Rodney Kinney, and Dan~S Weld.
\newblock S2orc: The semantic scholar open research corpus.
\newblock \emph{arXiv preprint arXiv:1911.02782}, 2019.

\bibitem[Su et~al.(2022)Su, Du, Yang, Zhou, Li, Rao, Sun, Lu, and Wen]{su2022MoMu}
Bing Su, Dazhao Du, Zhao Yang, Yujie Zhou, Jiangmeng Li, Anyi Rao, Hao Sun, Zhiwu Lu, and Ji-Rong Wen.
\newblock A molecular multimodal foundation model associating molecule graphs with natural language.
\newblock \emph{arXiv preprint arXiv:2209.05481}, 2022.

\bibitem[You et~al.(2020)You, Chen, Sui, Chen, Wang, and Shen]{you2020GraphCL}
Yuning You, Tianlong Chen, Yongduo Sui, Ting Chen, Zhangyang Wang, and Yang Shen.
\newblock Graph contrastive learning with augmentations.
\newblock \emph{Advances in neural information processing systems}, 33:\penalty0 5812--5823, 2020.

\bibitem[Beltagy et~al.(2019)Beltagy, Lo, and Cohan]{beltagy2019scibert}
Iz~Beltagy, Kyle Lo, and Arman Cohan.
\newblock Scibert: A pretrained language model for scientific text.
\newblock \emph{arXiv preprint arXiv:1903.10676}, 2019.

\bibitem[Liu et~al.(2023{\natexlab{b}})Liu, Nie, Wang, Lu, Qiao, Liu, Tang, Xiao, and Anandkumar]{liu2023MoleculeSTM}
Shengchao Liu, Weili Nie, Chengpeng Wang, Jiarui Lu, Zhuoran Qiao, Ling Liu, Jian Tang, Chaowei Xiao, and Animashree Anandkumar.
\newblock Multi-modal molecule structure--text model for text-based retrieval and editing.
\newblock \emph{Nature Machine Intelligence}, 5\penalty0 (12):\penalty0 1447--1457, 2023{\natexlab{b}}.

\bibitem[Das et~al.(2023)Das, Goyal, Lee, Bhattacharjee, and Ganguly]{das2023crysmmnet}
Kishalay Das, Pawan Goyal, Seung-Cheol Lee, Satadeep Bhattacharjee, and Niloy Ganguly.
\newblock {CrysMMNet}: Multimodal representation for crystal property prediction.
\newblock In Robin~J. Evans and Ilya Shpitser, editors, \emph{Proceedings of the Thirty-Ninth Conference on Uncertainty in Artificial Intelligence}, volume 216 of \emph{Proceedings of Machine Learning Research}, pages 507--517. PMLR, 31 Jul--04 Aug 2023.
\newblock URL \url{https://proceedings.mlr.press/v216/das23a.html}.

\bibitem[Ganose and Jain(2019)]{ganose2019robocrystallographer}
Alex~M Ganose and Anubhav Jain.
\newblock Robocrystallographer: automated crystal structure text descriptions and analysis.
\newblock \emph{MRS Communications}, 9\penalty0 (3):\penalty0 874--881, 2019.

\bibitem[Choudhary et~al.(2020)Choudhary, Garrity, Reid, DeCost, Biacchi, Hight~Walker, Trautt, Hattrick-Simpers, Kusne, Centrone, et~al.]{choudhary2020JARVIS}
Kamal Choudhary, Kevin~F Garrity, Andrew~CE Reid, Brian DeCost, Adam~J Biacchi, Angela~R Hight~Walker, Zachary Trautt, Jason Hattrick-Simpers, A~Gilad Kusne, Andrea Centrone, et~al.
\newblock The joint automated repository for various integrated simulations (jarvis) for data-driven materials design.
\newblock \emph{npj computational materials}, 6\penalty0 (1):\penalty0 173, 2020.

\bibitem[Sch{\"u}tt et~al.(2021)Sch{\"u}tt, Unke, and Gastegger]{schutt2021painn}
Kristof Sch{\"u}tt, Oliver Unke, and Michael Gastegger.
\newblock Equivariant message passing for the prediction of tensorial properties and molecular spectra.
\newblock In Marina Meila and Tong Zhang, editors, \emph{Proceedings of the 38th International Conference on Machine Learning}, volume 139 of \emph{Proceedings of Machine Learning Research}, pages 9377--9388. PMLR, 18--24 Jul 2021.
\newblock URL \url{https://proceedings.mlr.press/v139/schutt21a.html}.

\bibitem[Li et~al.(2021)Li, Liang, Zhao, Cui, Ouyang, Shao, Yu, and Yan]{li2021DeCLIP}
Yangguang Li, Feng Liang, Lichen Zhao, Yufeng Cui, Wanli Ouyang, Jing Shao, Fengwei Yu, and Junjie Yan.
\newblock Supervision exists everywhere: A data efficient contrastive language-image pre-training paradigm.
\newblock \emph{arXiv preprint arXiv:2110.05208}, 2021.

\bibitem[Shrivastava et~al.(2023)Shrivastava, Selvaraju, Naik, and Ordonez]{shrivastava2023clip-lite}
Aman Shrivastava, Ramprasaath~R. Selvaraju, Nikhil Naik, and Vicente Ordonez.
\newblock Clip-lite: Information efficient visual representation learning with language supervision.
\newblock In Francisco Ruiz, Jennifer Dy, and Jan-Willem van~de Meent, editors, \emph{Proceedings of The 26th International Conference on Artificial Intelligence and Statistics}, volume 206 of \emph{Proceedings of Machine Learning Research}, pages 8433--8447. PMLR, 25--27 Apr 2023.
\newblock URL \url{https://proceedings.mlr.press/v206/shrivastava23a.html}.

\bibitem[Gupta et~al.(2022)Gupta, Zaki, Krishnan, and Mausam]{gupta2022matscibert}
Tanishq Gupta, Mohd Zaki, NM~Anoop Krishnan, and Mausam.
\newblock Matscibert: A materials domain language model for text mining and information extraction.
\newblock \emph{npj Computational Materials}, 8\penalty0 (1):\penalty0 102, 2022.

\bibitem[Zeng et~al.(2022{\natexlab{b}})Zeng, Xiang, Yu, Wang, Li, Nussinov, and Cheng]{zeng2022ImageMol}
Xiangxiang Zeng, Hongxin Xiang, Linhui Yu, Jianmin Wang, Kenli Li, Ruth Nussinov, and Feixiong Cheng.
\newblock Accurate prediction of molecular properties and drug targets using a self-supervised image representation learning framework.
\newblock \emph{Nature Machine Intelligence}, 4\penalty0 (11):\penalty0 1004--1016, 2022{\natexlab{b}}.

\bibitem[Walker et~al.(2021)Walker, Trewartha, Huo, Lee, Cruse, Dagdelen, Dunn, Persson, Ceder, and Jain]{walker2021SOFC-slot}
Nicholas Walker, Amalie Trewartha, Haoyan Huo, Sanghoon Lee, Kevin Cruse, John Dagdelen, Alexander Dunn, Kristin Persson, Gerbrand Ceder, and Anubhav Jain.
\newblock The impact of domain-specific pre-training on named entity recognition tasks in materials science.
\newblock \emph{Available at SSRN 3950755}, 2021.

\bibitem[Venugopal et~al.(2021)Venugopal, Sahoo, Zaki, Agarwal, Gosvami, and Krishnan]{venugopal2021looking}
Vineeth Venugopal, Sourav Sahoo, Mohd Zaki, Manish Agarwal, Nitya~Nand Gosvami, and NM~Anoop Krishnan.
\newblock Looking through glass: Knowledge discovery from materials science literature using natural language processing.
\newblock \emph{Patterns}, 2\penalty0 (7), 2021.

\bibitem[McInnes et~al.(2018)McInnes, Healy, and Melville]{mcinnes2018umap}
Leland McInnes, John Healy, and James Melville.
\newblock Umap: Uniform manifold approximation and projection for dimension reduction.
\newblock \emph{arXiv preprint arXiv:1802.03426}, 2018.

\end{thebibliography}






\newpage
\section*{\textbf{Supplementary information}}
\renewcommand{\thefigure}{S\arabic{figure}}
\setcounter{figure}{0}
\setcounter{table}{0}
\setcounter{section}{0}
\section{GPT-synthesized narratives}

We introduce an example of GPT-synthesized narratives of KNiIO$_6$. These textual data were synthesized based on realistic properties from publicly available materials databases. This is a promising alternative to address the data scarcity and bias issues we currently face.

\begin{figure*}[h]
    \centering
    \includegraphics[width=0.9\textwidth]{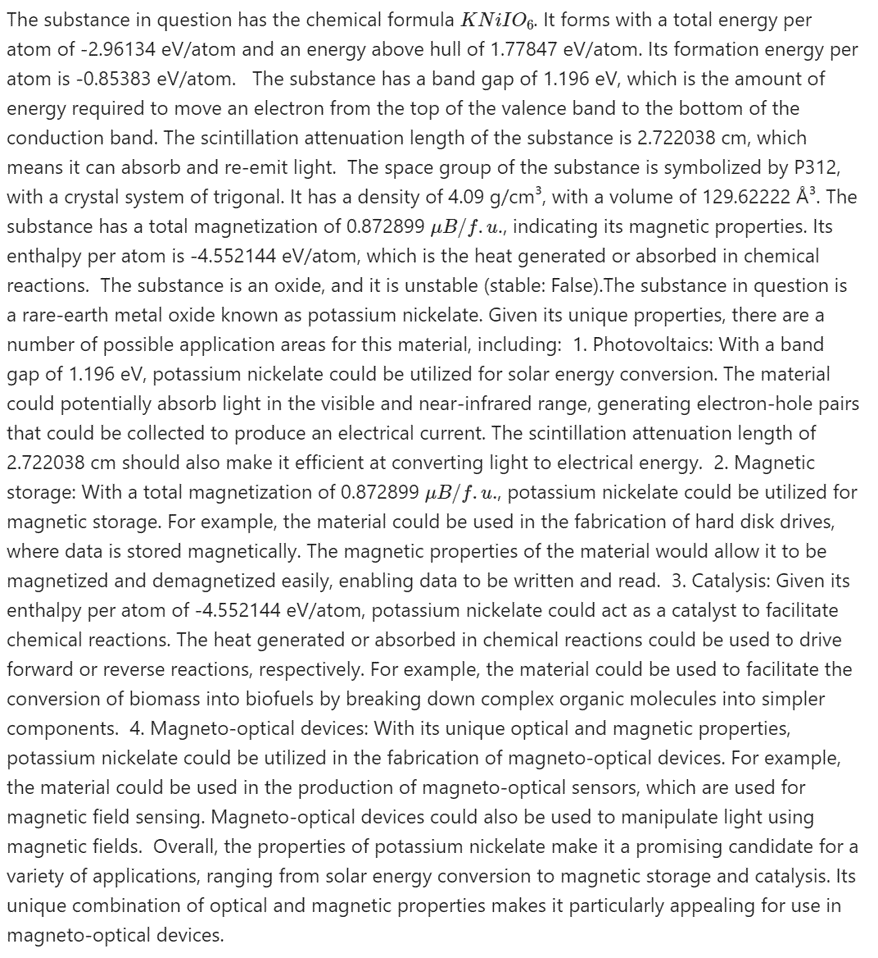}
    \caption{\textbf{Generated description about KNiIO$_6$}. The narrative describes the material property itself and infers potential applications of it.}
    \label{fig:example_narrative}
\end{figure*}

\mbox{~}
\clearpage
\newpage

\section{Zero-shot retrieval task}
Here we describe the evaluation process of the zero-shot retrieval task. There are two different ways according to the purpose: graph-to-text retrieval and text-to-graph retrieval. Graph-to-text retrieval is to find the closest text among the pool to the given materials graph as shown in Supplementary Fig. \ref{fig:retrieval_explanation}a. The query material and list of texts are encoded according to their corresponding encoders, and similarities between hidden representations are calculated. It is useful when we transfer knowledge to zero-shot classification tasks. On the other hand, Text-to-graph retrieval is to find the closest graph among the materials pool for the given query text as shown in Supplementary Fig. \ref{fig:retrieval_explanation}b. The list of materials and the query text are encoded according to their corresponding encoders, and similarities between hidden representations are calculated. 

\begin{figure*}[h]
    \centering
    \includegraphics[width=\textwidth]{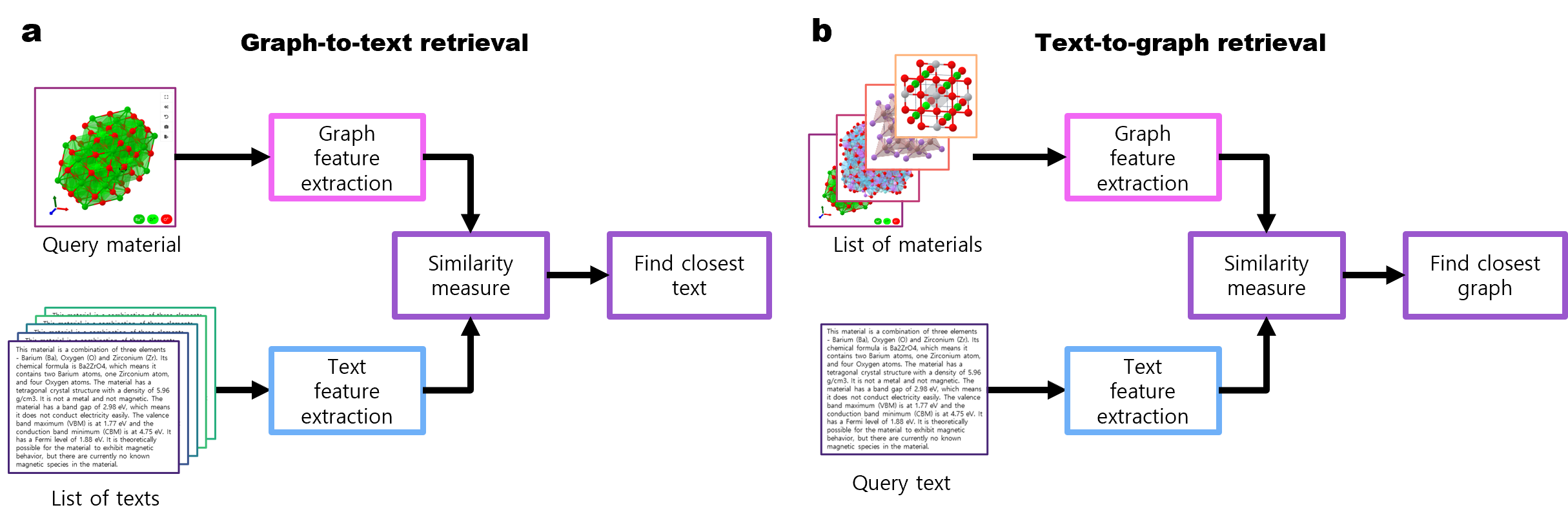}
    \caption{\textbf{Illustration of zero-shot retrieval process} Process of a) graph-to-text retrieval and b) text-to-graph retrieval. Two branches of the CLaC model extract features of corresponding entities and compare similarities to find the closest entity of the query.}
    \label{fig:retrieval_explanation}
\end{figure*}

\mbox{~}
\clearpage
\newpage

\section{Baseline models} \label{sec:baseline_models}
We designed 3 different baseline models to evaluate and compare the multimodal abilities of our CLaC model. 

\begin{figure*}[h]
    \centering
    \includegraphics[width=\textwidth]{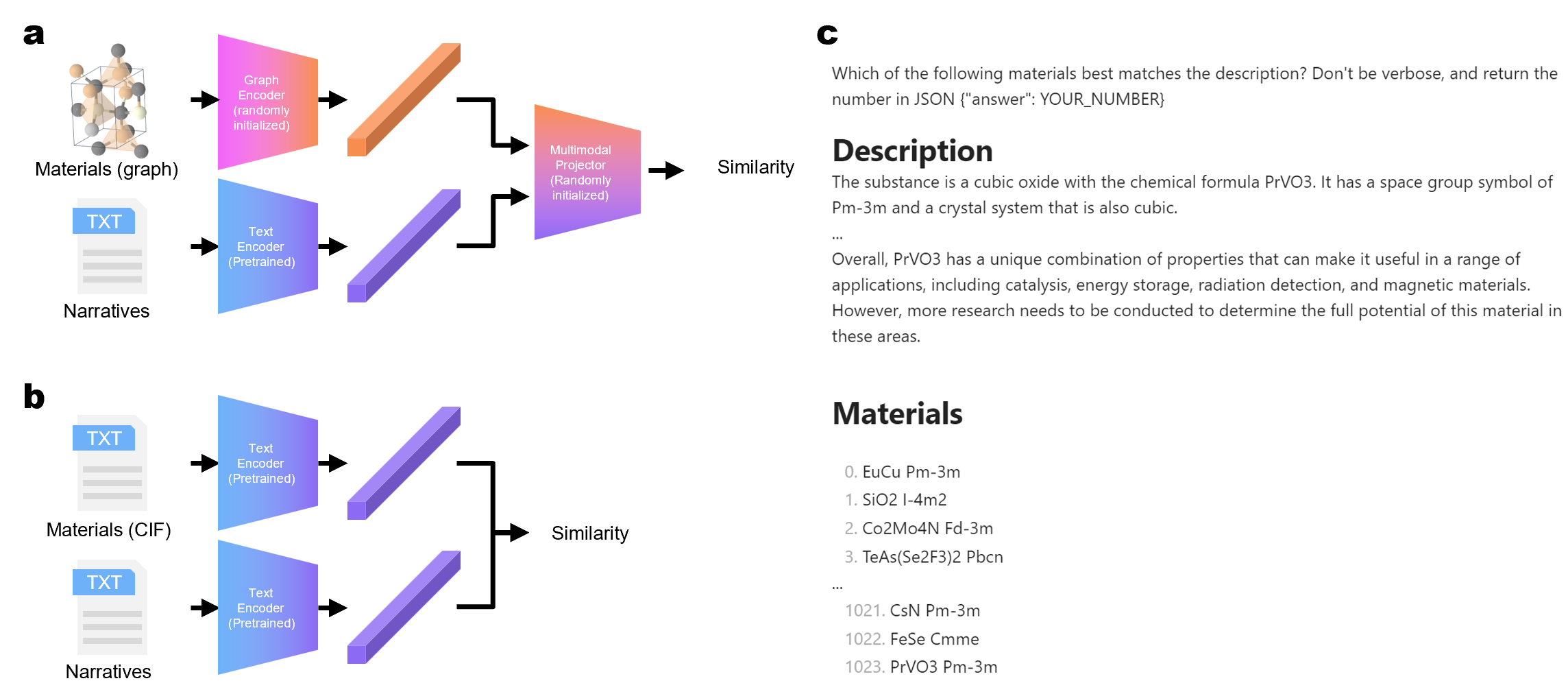}
    \caption{\textbf{Illustration of baseline models for zero-shot retrieval task} a) Our CLaC architecture with randomly initialized graph encoder and multimodal projector. b) The pre-trained BERT models extract features of materials and narratives and compare them. c) The narrative of the query materials and materials pool are provided in the prompt of GPT-style language model.}
    \label{fig:baseline_models}
\end{figure*}

The first baseline model is based on the same architecture as CLaC but the graph encoder and multimodal projector are randomly initialized (Supplementary Fig. \ref{fig:baseline_models}a. The second baseline model is pre-trained BERT models for the scientific domain, SciBERT and MatSciBERT. Similarities between hidden representations of CIF files and narratives are calculated and compared to select the closest one (Supplementary Fig. \ref{fig:baseline_models}b. The last baseline model is the GPT-style proprietary language model. The prompt is designed to select the closest material among the 1,024 materials pool as shown in Supplementary Fig. \ref{fig:baseline_models}c. The chemical formula and space group symbol of the materials pool are provided in the prompt.

First, we use the same architecture model as ours, including a randomly initialized graph encoder and randomly initialized multimodal projector to treat multimodality similarity calculation (baseline 1). We expect that accuracy would be less than 0.1\% when the baseline model predicts based on a randomly initialized graph encoder and multimodal projector. For the second baseline, we used a model with 2 branches that have pre-trained SciBERT or MatSciBERT as both text encoder and crystal encoder by inputting the crystallographic information file (CIF) of the corresponding crystal and calculating the cosine similarity between the two branches. Baseline 2 models are expected to find out the original pair by identifying information like the name of elements and crystal system semantics. The third baseline is in-context learning of ChatGPT-3.5-turbo. The given narrative of crystal and the pool of materials are provided in the prompt. For our models, We further evaluate and align the result according to the backbone model we used and the dataset on which the model was trained. 

\mbox{~}
\clearpage
\newpage




\section{Generating multiple-choice problems for zero-shot multimodal understanding evaluation}\label{generating_multiple_choice}

To evaluate the multimodal recognition performance of our model, we proposed multiple-choice problems for various subjects. To evaluate the compositional understanding of the developed model, we generate 10 sentences including only one correct answer. The correct answer is generated from the given materials while other incorrect answers are randomly synthesized as described in Supplementary Fig. \ref{fig:multiple_choice_examples}.

\begin{figure*}[h]
    \centering
    \includegraphics[width=\textwidth]{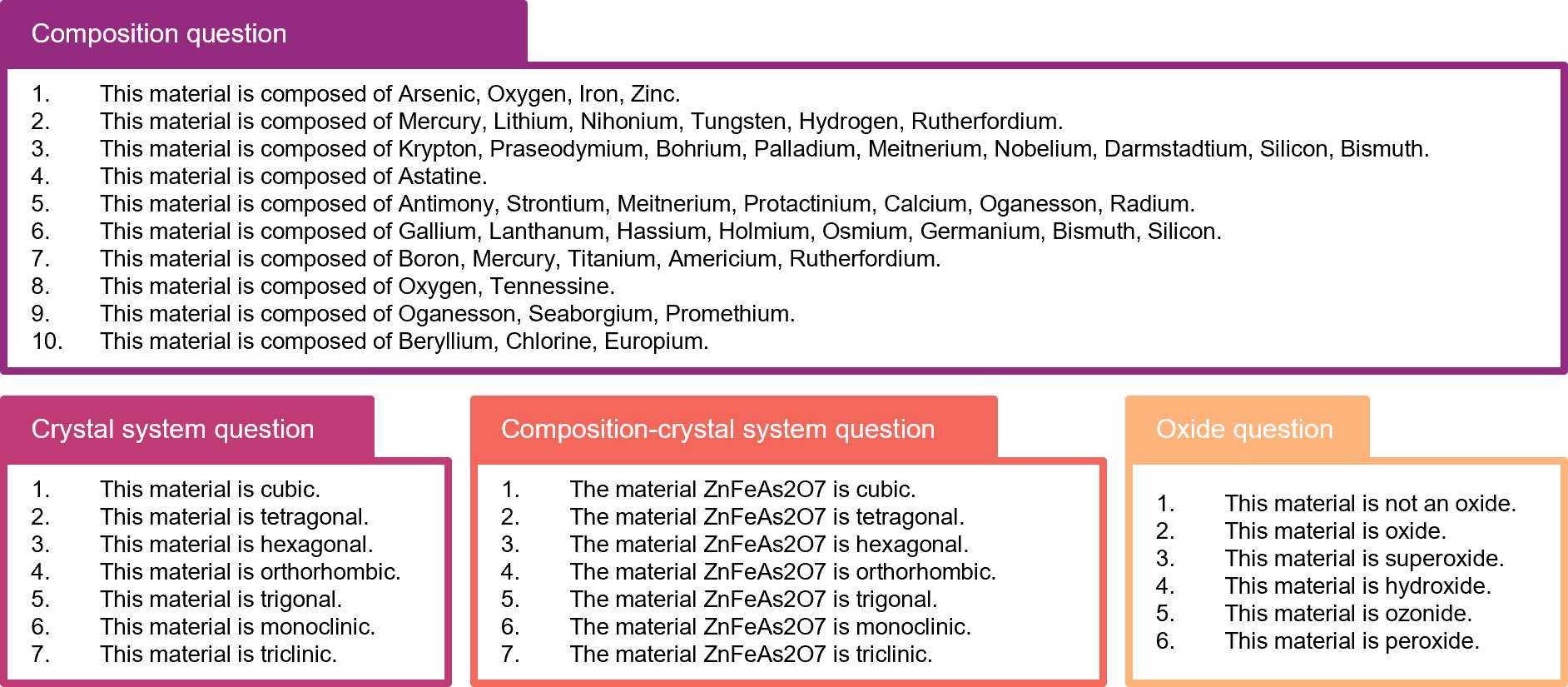}
    \caption{\textbf{Examples of multiple choice problems for evaluating various understandings of the CLaC model.} Multiple-choice problems are designed to evaluate the basic zero-shot understanding for materials science.}
    \label{fig:multiple_choice_examples}
\end{figure*}

\mbox{~}
\clearpage
\newpage

\section{Visualization of latent space according to keywords}\label{latent_keywords}

We analyzed and visualized crystal-text pair data by labeling each text entry based on the presence or absence of specific keywords. The results revealed that data with similar labels exhibited clustering within the latent space. This observation suggests a promising new approach for materials discovery leveraging linguistic analysis.

\begin{figure*}[h]
    \centering
    \includegraphics[width=\textwidth]{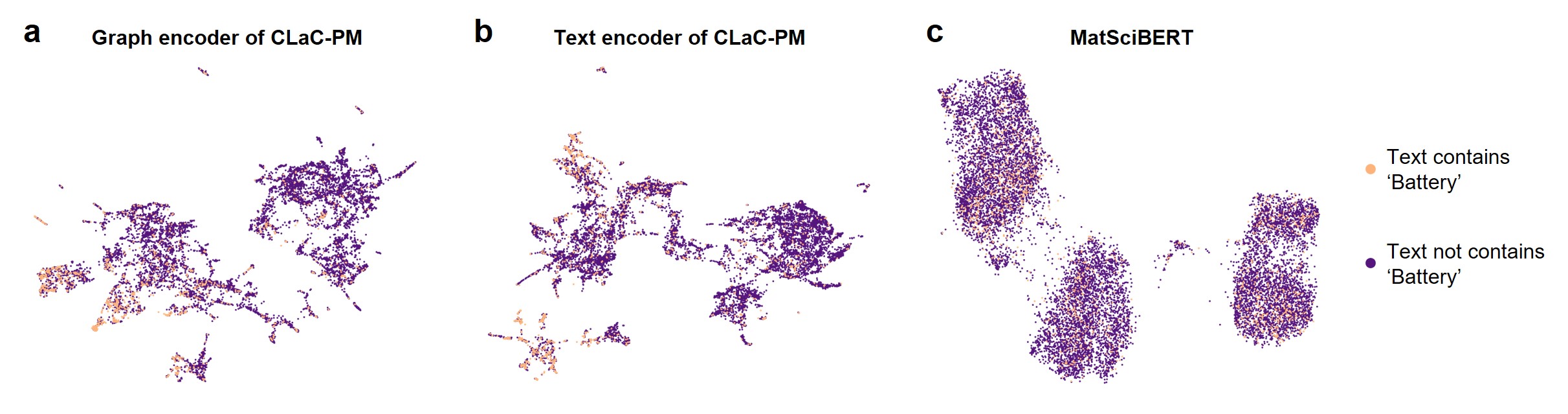}
    \caption{\textbf{UMAP visualization of graph and text representations.} Each column corresponds to graph representation (\textbf{a, d, g, j}), text representation after multimodal training (\textbf{b, e, h, k}), and text representation of pristine MatSciBERT (\textbf{c, f, i, l}). Each row corresponds to oxide type classification (\textbf{a, b, c}), crystal system classification (\textbf{d, e, f}), classification of the text including `battery' (\textbf{g, h, i}), and classification of the text including `fuel cell' (\textbf{j, k, l}).}
    \label{fig:embedding_battery}
\end{figure*}

\mbox{~}
\clearpage
\newpage

\begin{figure*}[h]
    \centering
    \includegraphics[width=\textwidth]{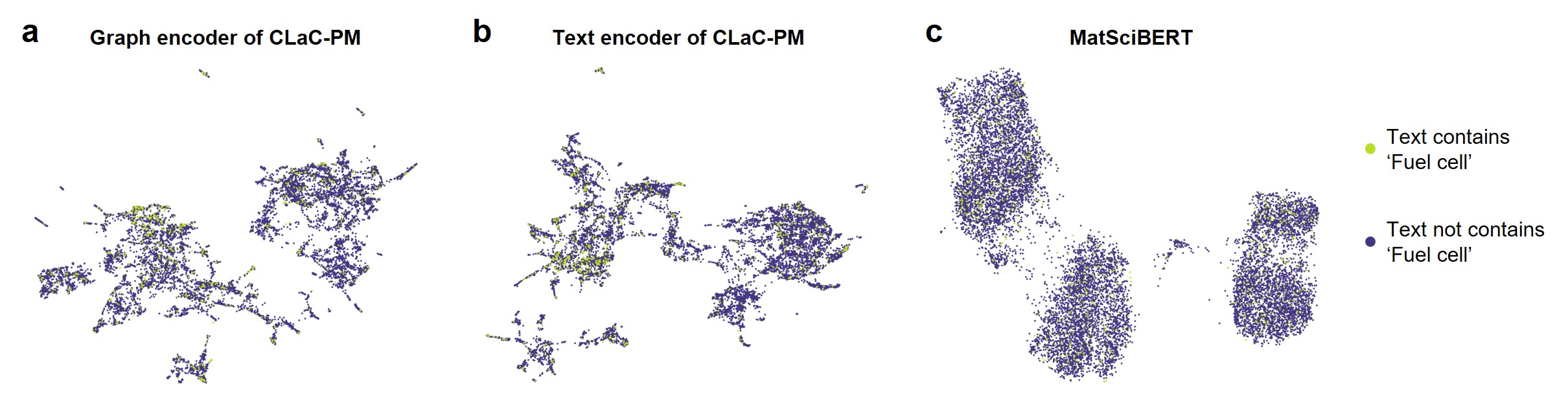}
    \caption{\textbf{UMAP visualization of graph and text representations.} Each column corresponds to graph representation (\textbf{a, d, g, j}), text representation after multimodal training (\textbf{b, e, h, k}), and text representation of pristine MatSciBERT (\textbf{c, f, i, l}). Each row corresponds to oxide type classification (\textbf{a, b, c}), crystal system classification (\textbf{d, e, f}), classification of the text including `battery' (\textbf{g, h, i}), and classification of the text including `fuel cell' (\textbf{j, k, l}).}
    \label{fig:embedding_fuelcell}
\end{figure*}

\mbox{~}
\clearpage
\newpage

\begin{figure*}[h]
    \centering
    \includegraphics[width=\textwidth]{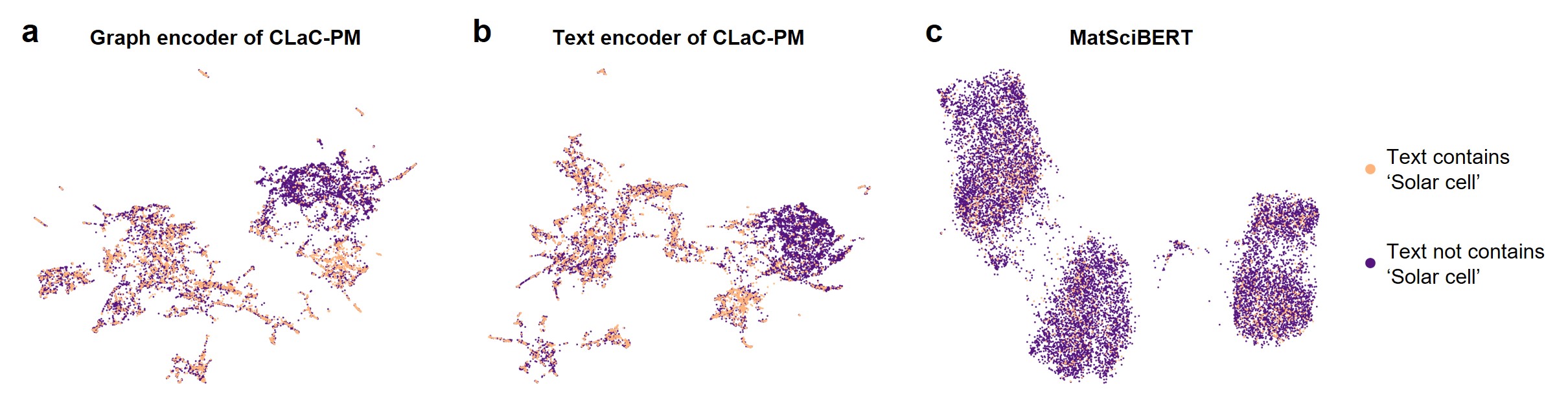}
    \caption{\textbf{UMAP visualization of graph and text representations.} Each column corresponds to graph representation (\textbf{a, d, g, j}), text representation after multimodal training (\textbf{b, e, h, k}), and text representation of pristine MatSciBERT (\textbf{c, f, i, l}). Each row corresponds to oxide type classification (\textbf{a, b, c}), crystal system classification (\textbf{d, e, f}), classification of the text including `battery' (\textbf{g, h, i}), and classification of the text including `fuel cell' (\textbf{j, k, l}).}
    \label{fig:embedding_solarcell}
\end{figure*}

\mbox{~}
\clearpage
\newpage

\begin{figure*}[h]
    \centering
    \includegraphics[width=\textwidth]{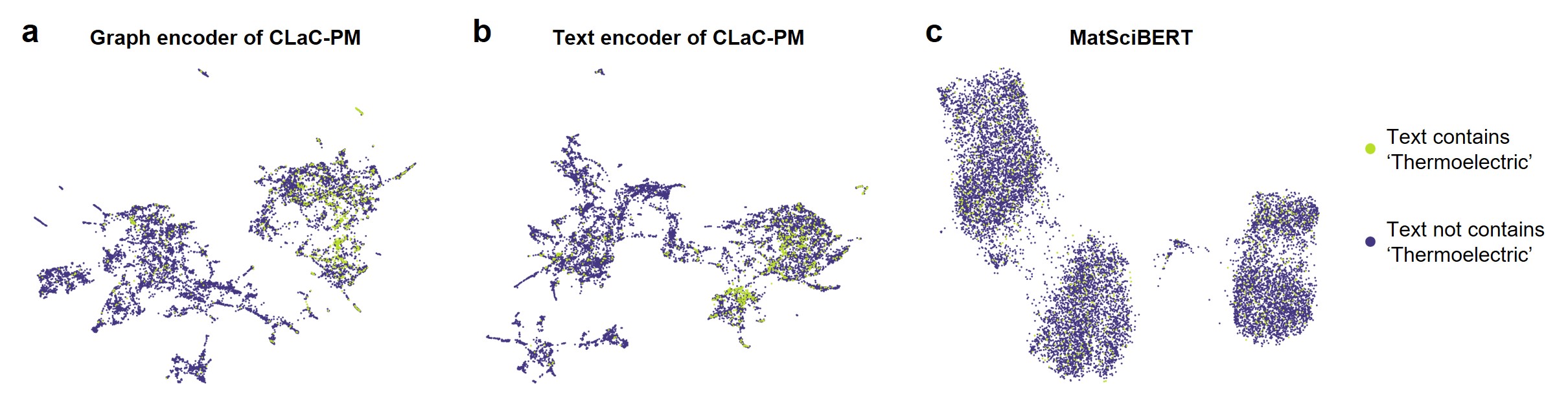}
    \caption{\textbf{UMAP visualization of graph and text representations.} Each column corresponds to graph representation (\textbf{a, d, g, j}), text representation after multimodal training (\textbf{b, e, h, k}), and text representation of pristine MatSciBERT (\textbf{c, f, i, l}). Each row corresponds to oxide type classification (\textbf{a, b, c}), crystal system classification (\textbf{d, e, f}), classification of the text including `battery' (\textbf{g, h, i}), and classification of the text including `fuel cell' (\textbf{j, k, l}).}
    \label{fig:embedding_thermoelectric}
\end{figure*}
\end{document}